\documentclass[10pt]{article} 
\usepackage[preprint]{tmlr}

\usepackage{amsmath,amsfonts,bm}

\def\eqref#1{equation~\ref{#1}}

\def\1{\bm{1}}

\DeclareMathAlphabet{\mathsfit}{\encodingdefault}{\sfdefault}{m}{sl}
\SetMathAlphabet{\mathsfit}{bold}{\encodingdefault}{\sfdefault}{bx}{n}

\newcommand{\E}{\mathbb{E}}

\newcommand{\R}{\mathbb{R}}

\newcommand{\KL}{D_{\mathrm{KL}}}
\newcommand{\Var}{\mathrm{Var}}

\DeclareMathOperator{\sign}{sign}

\usepackage{hyperref}
\usepackage{url,scalerel,stackengine}
\stackMath
\newcommand\reallywidehat[1]{%
\savestack{\tmpbox}{\stretchto{%
  \scaleto{%
    \scalerel*[\widthof{\ensuremath{#1}}]{\kern-.6pt\bigwedge\kern-.6pt}%
    {\rule[-\textheight/2]{1ex}{\textheight}}
  }{\textheight}%
}{0.5ex}}%
\stackon[1pt]{#1}{\tmpbox}%
}
\usepackage{enumitem}
\usepackage{amsmath,amssymb}
\usepackage{bbm}
\usepackage{caption}
\usepackage{wrapfig,xcolor,tcolorbox}
\usepackage{algorithm,algpseudocode}
\usepackage{subfigure}
\usepackage{tikz}
\usetikzlibrary{decorations.pathreplacing}

\usepackage{array}
\newcolumntype{C}[1]{>{\centering\let\newline\\\arraybackslash\hspace{0pt}}m{#1}}

\newcommand{\cK}{\mathcal{K}}

\newcommand{\cN}{\mathcal{N}}

\newcommand{\cP}{\mathcal{P}}

\newcommand{\kD}{\mathfrak{D}}

\newcommand{\kR}{\mathfrak{R}}

\usepackage[nohints]{minitoc}

\renewcommand{\P}{\mathbb{P}}
\newcommand{\N}{\mathbb{N}}
\newcommand{\Hub}{\mathrm{Hub}}
\newcommand{\SHub}{\mathrm{SeqHub}}
\renewcommand{\eqref}[1]{(\ref{#1})}

\renewcommand{\d}{\mathrm{d}}
\usepackage{todonotes}

\newcommand{\HubUCB}{{\color{red!50!black}\texttt{HuberUCB}}}
\newcommand{\SeqHubUCB}{{\color{red!50!black}\texttt{SeqHuberUCB}}}

\newcommand{\DeltaEpsi}{\widetilde{\Delta}_{i,\varepsilon}}

\newtheorem{Theorem}{Theorem}
\newtheorem{Corollary}{Corollary}
\newtheorem{Lemma}{Lemma}
\newtheorem{remark}{Remark}
\newtheorem{definition}{Definition}

\newlist{npf}{enumerate}{2}
\setlist[npf,1]{label={\textbf{Step \arabic*.}}, ref={Step \arabic*},leftmargin=*, itemindent=3.5em}
\RequirePackage{changepage}
\newcommand{\pf}{\textsc{Proof}}
\newcommand{\onepf}[1]{\linebreak[0]\pf : #1 }

\title{Bandits Corrupted by Nature:\\ Lower Bounds on Regret and Robust Optimistic Algorithms}

\author{\name Debabrota Basu \email debabrota.basu@inria.fr
	\AND
	\name Odalric-Ambrym Maillard \email odalric.maillard@inria.fr
	\AND
	\name Timothée Mathieu \email timothee.mathieu@inria.fr \AND
	\addr Université de Lille, Inria, CNRS, Centrale Lille
	UMR 9189 – CRIStAL, F-59000 Lille, France}

\def\month{MM}  
\def\year{YYYY} 
\def\openreview{\url{https://openreview.net/forum?id=XXXX}} 

\begin{document}

\maketitle

\doparttoc
\faketableofcontents

	\begin{abstract}%
	We study the corrupted bandit problem, i.e. a stochastic multi-armed bandit problem with $k$ unknown reward distributions, which are heavy-tailed and corrupted by a history-independent adversary or Nature. To be specific, the reward obtained by playing an arm comes from corresponding heavy-tailed reward distribution with probability $1-\varepsilon \in (0.5,1]$ and an arbitrary corruption distribution of unbounded support with probability $\varepsilon \in [0,0.5)$.
	First, we provide \textit{a problem-dependent lower bound on the regret} of any corrupted bandit algorithm. The lower bounds indicate that the corrupted bandit problem is harder than the classical stochastic bandit problem with sub-Gaussian or heavy-tail rewards.
	Following that, we propose a novel UCB-type algorithm for corrupted bandits, namely \HubUCB, that builds on Huber's estimator for robust mean estimation. Leveraging a novel concentration inequality of Huber's estimator, we prove that \HubUCB\ achieves a near-optimal regret upper bound.
	Since computing Huber's estimator has quadratic complexity, we further introduce a sequential version of Huber's estimator that exhibits linear complexity. We leverage this sequential estimator to design \SeqHubUCB\ that enjoys similar regret guarantees while reducing the computational burden.
	Finally, we experimentally illustrate the efficiency of \HubUCB\ and \SeqHubUCB\ in solving corrupted bandits for different reward distributions and different levels of corruptions.
	\end{abstract}

\section{Introduction}
Multi-armed bandit problem is an archetypal setting to study sequential decision-making under incomplete information~\citep{lattimore2018bandit}.
In the classical setting of stochastic multi-armed bandits, the decision maker or agent has access to $k\in\mathbb{N}$ unknown reward distributions or arms.
At every step, the agent plays an arm and obtains a reward. The goal of the agent is to maximize the expected total reward accumulated by a given horizon $T\in\mathbb{N}$.

In this paper, we are interested in a challenging extension of the classical multi-armed bandit problem, where the reward at each step is corrupted by Nature, which is a stationary mechanism independent of the agent's decisions and observations. This setting is often referred as the \textit{Corrupted Bandits}.
Specifically, we extend the existing studies of corrupted bandits~\citep{lykouris2018stochastic,bogunovic2020corruption,kapoor2019corruption} to the more general case, where the `true' reward distribution might be heavy-tailed (i.e. with a finite number of finite moments) and the corruption can be unbounded.

\paragraph{A Motivating Example: Treatments of Varroa Mites.}
Though this article focuses on the theoretical aspects of this problem, we hereby illustrate a case study with roots in agriculture that motivates us.
A bee-keeper has to choose between a set of treatments to save her bees from varroa mites.
Every year, the bee-keeper must rotate between the treatments as the varroa mites develop resistance to a given treatment~\citep{rinkevich2020detection,kamler2016comparison}.
The goal of the bee-keeper is to choose a sequence of treatments over the years that eliminate as many number of varroa mites as possible.
The reward of a treatment is measured by the number of fallen varroa mites due to it.
This reward function is heavy-tailed.
As the number of fallen varroa mites is counted manually by the bee-keeper, this process is prone to human error.

The corruption in particular has been witnessed empirically while plotting the number of fallen mites for a given treatment (Fig. 1~\citep{amitrazcomparison}). To understand the phenomena of heavy-tailedness and corruption, let us imagine a natural model for the number of dead mites in a given interval of 1 
week, i.e. a Poisson 
distribution (light-tailed distribution). In this case, if the mean of the Poisson distribution is $\lambda=100$, which is the case in 
Fig. 1~\citep{amitrazcomparison}. Then, the standard deviation of the number of dead mites should be equal 
to $\sqrt{100}=10$, which contrasts with the standard deviation larger than $200$
as reported in the Fig. 1~\citep{amitrazcomparison}. Now, it is hard to 
decouple why this heavy-tailedness suddenly appear, and if there are 
outliers, what is their distribution or the upper bound on them. Thus, to deal with such observations of reward, we aim to design algorithms that can 
deal with both heavy-tailedness and unbounded corruptions.

Interestingly, in this problem, corruptions in the measured rewards are natural and non-adversarial but possibly unbounded. The heavy-tailed and corrupted nature of the problem resists application of the non-robust bandit algorithms, such as UCB~\citep{auer2002finite}, and motivates us to introduce the setting of \textit{Bandits corrupted by Nature}.

\paragraph{Bandits corrupted by Nature.} Motivated by the aforementioned example, we model a corrupted reward distribution as $(1-\varepsilon)P+\varepsilon H$, where $P$ is the distribution of inliers with a finite variance, $H$ is the distributions of outliers with probably unbounded support, and $\varepsilon \in [0,1/2)$ is the proportion of outliers.
Thus, in the corresponding stochastic bandit setting, an agent has access to $k$ arms of corrupted reward distributions $\lbrace (1-\varepsilon)P_i+\varepsilon H_i\rbrace_{i=1}^k$. Here, $P_i$'s are uncorrupted reward distributions with heavy-tails and bounded variances, and $H_i$'s are corruption distributions with probably unbounded corruptions. The goal of the agent is to maximize the expected total reward accumulated oblivious to the corruptions. 
This is equivalent to considering a setting where at every step Nature flips a coin with success probability $\varepsilon$. The agent obtains a corrupted reward if Nature obtains $1$ and otherwise, an uncorrupted reward. We call this setting \textit{Bandits corrupted by Nature} as the corruption introduced in each step does not depend on the present or previous choices of arms and observed rewards.
Our setting encompasses both heavy-tailed rewards and unbounded corruptions. We formally define the setting and corresponding regret definition in Section~\ref{sec:setting}.

Bandits corrupted by Nature is different from the adversarial bandit setting~\citep{auer2002nonstochastic}.
The adversarial bandit assumes existence of a non-stochastic adversary that can return at each step the worst-case reward to the agent depending on its history of choices.
Incorporating corruptions in this setting, \cite{lykouris2018stochastic,bogunovic2020corruption} consider settings where the rewards can be corrupted by a history-dependent adversary but the total amount of corruption and also the corruptions at each step are bounded.
However, we encounter problems in ecology and agronomy, such as treatments against varroa mites, where the corruptions are not adversarial, and are independent of the previous history of decisions.
Thus, in contrast to the adversarial corruption setting in literature, we consider a non-adversarial proportion of corruptions ($\varepsilon \in [0,1/2)$) at each step, which are stochastically generated from unbounded corruption distributions $\left(\lbrace H_i\rbrace_{i=1}^k\right)$.
To the best of our knowledge, only \cite{kapoor2019corruption} have studied similar non-adversarial corruption setting with a history-independent proportion of corruption at each step.
But they assume that the probable corruptions at each step are bounded, and the uncorrupted rewards are sub-Gaussian.
Hence, we observe that \textit{there is a gap in the literature in studying unbounded stochastic corruption for bandits with probably heavy-tailed rewards and this article aims to fill this gap.}
Specifically, we aim to deal with unbounded corruption and heavy-tails simultaneously, which requires us to develop a novel sensitivity analysis of the robust estimator in lieu of a worst-case (adversarial bandits) analysis.

\noindent\textbf{Our Contributions.} Specifically, in this paper, we aim to investigate three main questions:
\begin{tcolorbox}
        \begin{enumerate}[nosep,leftmargin=*]
        \item Is the setting of bandits corrupted by Nature with unbounded corruptions and heavy tails fundamentally harder (in terms of the regret lower bound) than the classical sub-Gaussian and uncorrupted bandit setting?
        \item Is it possible to design an \textit{efficient and robust algorithm} that achieves an order-optimal performance (\textit{logarithmic} regret) in the corrupted by Nature setting?
        \item Are robust bandit algorithms \textit{efficient in practice}?
        \end{enumerate}
\end{tcolorbox}
\noindent These questions have led us to the following contributions:

\noindent \textit{1. Hardness of bandits corrupted by Nature with unbounded corruptions and heavy tails.} In order to understand the fundamental hardness of the proposed setting, we use a suitable notion of regret, denoted by $\kR_n$, (Equation~\eqref{eq:def_regret},~\citep{kapoor2019corruption}) that extends the traditional pseudo-regret~\citep{lattimore2018bandit} to the corrupted setting. Then, in Section~\ref{sec:lower_bounds}, we derive lower bounds on regret that reveal increased difficulties of corrupted bandits with heavy tails in comparison with the classical non-corrupted and light-tailed Bandits. (a) In the Heavy-tailed regime (\ref{lem:bound_KL_student}), we show that even when the suboptimality gap $\Delta_i$\footnote{The suboptimality gap of an arm is the difference in mean rewards of an optimal arm and that arm.} is large, the regret increase with $\Delta_i$ because of the difficulty to distinguish between two arms when the rewards of Heavy-tailed.
(b) Our lower bounds 
indicate that when $\Delta_i$ is large, the logarithmic regret is asymptotically achievable, but the hardness depends on the corruption proportion $\varepsilon$, variance of $P_i$, i.e. $\sigma_i$, and the suboptimality gap $\Delta_i$.
Specifically, if $\frac{\Delta_i}{\sigma_i}$'s are small, i.e. we are in low distinguishability/high variance regime, the hardness is dictated by $\frac{\sigma_i^2}{\overline{\Delta}_{i,\varepsilon}^2}$. Here, $\overline{\Delta}_{i,\varepsilon} \triangleq \Delta_i(1-\varepsilon)  - 2\varepsilon \sigma_i$ is the `\textit{corrupted suboptimality gap}' that replaces the traditional suboptimality gap $\Delta_i$ in the lower bound of non-corrupted and light-tailed bandits~\citep{lairobbins1985}. Since $\overline{\Delta}_{i,\varepsilon} \leq {\Delta}_{i}$, it is harder to distinguish the optimal and suboptimal arms in the corrupted settings. They are the same when the corruption proportion $\varepsilon=0$.

Additionally, our analysis addresses an open problem in heavy-tailed bandits. Works on heavy-tailed bandits \citep{Bubeck2013BanditsWH,agrawal2021regret} rely on the assumption that a bound on the $(1\!+\!\varepsilon)$-moment, i.e. $\E[|X|^{1+\varepsilon}]$, is known for some $\varepsilon>0$. We do not assume such a restrictive bound as knowing a bound on $\E[|X|^{1+\varepsilon}]$ implies the knowledge of a bound on the sub-optimality gap $\Delta$. Instead, we assume that the centered moment, specifically the variance, is bounded by a known constant. Thus, we address the open problem mentioned in~\citep{agrawal2021regret} by relaxing the classical bounded $(1\!+\!\varepsilon)$-moment assumption with the bounded centered moment one.

\noindent	\textit{2. Robust and Efficient Algorithm Design.} In Section~\ref{sec:upper_bound}, we propose a robust algorithm, called \HubUCB, that leverages the Huber's estimator for robust mean estimation. We derive a novel concentration inequality on the deviation of empirical Huber's estimate that allows us to design robust and tight confidence intervals for \HubUCB.
In Theorem~\ref{th:upper_bound}, we show that \HubUCB\ achieves the logarithmic regret, and also the optimal rate when the sub-optimality gap $\Delta$ is not too large.  We show that for \HubUCB, $\kR_n$ can be decomposed according to the respective values of $\Delta_i$ and $\sigma_i$:
\begin{align*}
\kR_n~~\le~~\underbrace{\mathcal{O}\left(\sum_{i : \Delta_i > \sigma_i}\log(n)\sigma_i\right)}_{\text{Error due to Heavy-tail}} + \underbrace{\mathcal{O}\left(\sum_{i:\Delta_i \le \sigma_i} \log(n)\Delta_i\frac{\sigma_i^2}{\overline{\Delta}_{i,\varepsilon}^2}\right)}_{\text{$\sigma^2/\Delta$ error with corrupted sub-optimality gaps}}.
\end{align*}
Thus, our upper bound allows us to segregate the errors due to heavy-tail, corruption, and corruption-correction with heavy tails. The error incurred by \HubUCB\ can be directly compared to the lower bounds obtained in Section~\ref{sec:lower_bounds} and interpreted in both the high distinguishibility regime and the low distinguishibility regime as previously mentioned.

\noindent	\textit{3. Empirically Efficient and Robust Performance.} To the best of our knowledge, we present the first robust mean estimator that can be computed in a linear time in a sequential setting (Section~\ref{sec:sequentialhub}). Existing robust mean estimators, such as Huber's estimator, need to be recomputed at each iteration using all the data, which implies a quadratic complexity. Our proposal recomputes Huber's estimator only when the iteration number is a power of $2$ and computes a sequential approximation on the other iterations. We use the Sequential Huber's estimator to propose \SeqHubUCB. We theoretically show that \SeqHubUCB\ achieves similar order of regret as \HubUCB, while being computationally efficient. 
In Section~\ref{sec:examples}, we also experimentally illustrate that \HubUCB\ and \SeqHubUCB\ achieve the claimed performances for corrupted Gaussian and Pareto environments.

We further elaborate on the novelty of our results and position them in the existing literature in Section~\ref{sec:related}. For brevity, we defer the detailed proofs and the parameter tuning to Appendix.

\section{Related Work}\label{sec:related}
Due to the generality of our setting, this work either extends or relates to the existing approaches in both the heavy-tailed and corrupted bandits literature. While designing the algorithm, we further leverage the literature of robust mean estimation. In this section, we connect to these three streams of literature. Table~\ref{fig:table_lit} summarizes the previous works and posits our work in lieu.

\begin{table}[ht!]
\begin{center}
\begin{footnotesize}
\begin{tabular}{|C{12em}|C{6em}|c|c|c|C{6em}|}
  \hline
  Algorithms & Settings & Corruption & Type of outliers & Heavy-tailed & Adversarial/\\
   & & & & & Stochastic\\
  \hline
\hline
Our work & MAB & Yes & Unbounded & Yes &  Stochastic\\
  \hline
\cite{Bubeck2013BanditsWH,agrawal2021regret,lee2020optimal} & MAB & No & x & Yes &  Stochastic\\
\hline
\cite{lykouris2018stochastic}  & MAB & Yes & Bounded & No &  Stochastic\\
  \hline
\cite{bogunovic2020corruption} & GP Bandits & Yes & Bounded & No & Adversarial\\
\hline
\cite{kapoor2019corruption} & MAB \& Linear Bandits & Yes & Bounded & No &  Stochastic\\
  \hline
\cite{medina2016no,shao2018almost} & Linear Bandits & No & x & Yes &  Stochastic\\

  \hline
  \cite{bouneffouf2021corrupted} & Contextual Bandits &  context only & Unbounded &  No & Stochastic\\
  \hline
  \cite{pmlr-v97-agarwal19c} & Control & Yes & Bounded & x &  Adversarial\\
  \hline
  \cite{hajiesmaili2020adversarial,auer2002nonstochastic,pogodin2020first}  & MAB & Yes & Bounded & x &  Adversarial\\
  \hline
\end{tabular}
\end{footnotesize}
\caption{Comparison of existing results on Corrupted and Heavy-tailed Bandits.\label{fig:table_lit}}
\end{center}\vspace*{-1.5em}
\end{table}

\noindent\textit{Heavy-tailed bandits.} \cite{Bubeck2013BanditsWH} are one of the first to study robustness in multi-armed bandits by studying the heavy-tailed rewards. They use robust mean estimator to propose the RobustUCB algorithms. They show that under assumptions on the raw moments of the reward distributions, a logarithmic regret is achievable. It sprouted research works leading to either tighter rates of convergence~\citep{lee2020optimal,agrawal2021regret}, or algorithms for structured environments~\citep{medina2016no,shao2018almost}. Our article uses Huber's estimator which was already discussed in \citep{Bubeck2013BanditsWH}. However, the chosen parameters in~\citep{Bubeck2013BanditsWH} were suited for heavy-tailed distributions, and thus, \textit{render their proposed estimator non-robust to corruption. We address this gap in this work.}

\noindent\textit{Corrupted bandits.} The existing works on Corrupted Bandits~\citep{lykouris2018stochastic,bogunovic2020corruption,kapoor2019corruption} are restricted to bounded corruption. When dealing with bounded corruption, one can use techniques similar to adversarial bandits~\cite{auer2002nonstochastic} to deal with an adversary that can't corrupt an arm too much. The algorithms and proof techniques are fundamentally different in our article because the stochastic (or non-adversarial) corruption by Nature allows us to learn about the inlier distribution on the condition that corresponding estimators are robust. Thus, \textit{our bounds retain the problem-dependent regret, while successfully handling probably unbounded corruptions with robust estimators}.

\noindent\textit{Robust mean estimation}.
 Our algorithm design leverages the rich literature of robust mean estimation, specifically the influence function representation of Huber's estimator. The problem of robust mean estimation in a corrupted and heavy-tailed setting stems from the work of Huber~\citep{Huber1964RobustEO, huber2004robust}. Recently, in tandem with machine learning, there have been numerous advances both in the heavy-tailed~\citep{devroye2016sub,catoni2012challenging,minsker2019distributed}, and in the corrupted settings~\citep{lecue2020robust,minsker2021robust,prasad2019unified, prasad2020robust,depersin2019robust,lerasle2019monk,lecue2020robust}. Our work, specifically the novel concentration inequality for Huber's estimator, enriches this line of work with a result of parallel interest. We introduce a sequential version of Huber's estimator achieving linear complexity.

	\section{Bandits corrupted by Nature: Problem formulation}\label{sec:setting}
	In this section, 
	we present the corrupted bandits setting that we study,  together with the corresponding notion of regret.
	Similarly to the classical bandit setup, the regret decomposition lemma allows us to focus on the expected number of pulls of a suboptimal arm as the central quantity to control algorithmic standpoint.
	
	\noindent\textbf{Notations.}
We denote by $\mathcal{P}$ the set of probability distributions on the real line $\R$ and by $\mathcal{P}_{[q]}\triangleq\{ P \in \mathcal{P} : \E_P[|X|^q]<\infty\}$ the set of distributions with at least $q \geq 1$ finite moments. $\1\{A\}$ is the indicator function for the event $A$ being true. We denote the mean of a distribution $P_i$ as $\mu_i \triangleq \E_{P_i}[X]$. For any $\mathcal{D}\subset \mathcal{P}$, we denote $\mathcal{D}(\varepsilon)\triangleq \{(1-\varepsilon)P+\varepsilon H: \, P \in \mathcal{D}, H \in \mathcal{P}\}$ the set of corrupted distributions from $\mathcal{D}$.

	\paragraph{Problem Formulation.}
	In the setting of \textit{Bandits corrupted by Nature}, a bandit algorithm faces an environment with $k\in \mathbb{N}$ many reward distributions in the form
	$\lbrace (1-\varepsilon)P_i+\varepsilon H_i\rbrace_{i=1}^k$.
	Here $P_i,H_i$ are real-valued distributions and $\varepsilon$ is a mixture parameter assumed to be in $[0,1/2)$, that is
	$P_i$ is given more weights than $H_i$ in the mixture of arm $i$. For this reason the $\lbrace P_i\rbrace_{i=1}^k$ are called the \emph{inlier} distributions and the  $\lbrace H_i\rbrace_{i=1}^k$ the \emph{outlier} distributions.
	We assume the inlier distributions have at least $2$ finite moments that is $P_1,\dots,P_k\!\in\! \mathcal{P}_{[2]}$,
	while no restriction is put on the outlier distributions, that is $H_1,\dots,H_k\!\in\! \mathcal{P}$.
	For this reason, we also refer to the outlier distributions as the \emph{corrupted} distributions, and to the inlier distributions as the \emph{non-corrupted} ones.
	$\varepsilon$ is called the level of corruption. We write $\nu^\varepsilon$ the law of the corrupted environment, and we refer to that of the non-corrupted environment  $\nu^0$ by $\nu$.

	The game proceed as follows:
	At each step $t \in \lbrace 0, \ldots, n\rbrace$, the agent policy $\pi$ interacts with the corrupted environment by choosing an arm $A_t$ and obtaining a reward corrupted by Nature.
	To generate this reward, Nature first draws a random variable $C_t \in \{0,1\}$ from a Bernoulli distribution with mean $\varepsilon \in [0,1/2)$. If $C_t=1$, it generates a corrupted reward $Z_t$ from distribution $H_{A_t}$ corresponding to the chosen arm $A_t \in \{1,\ldots, k\}$. Otherwise, it generates a non-corrupted $X_t'$ from distribution  $P_{A_t}$. More formally, Nature generates reward $X_t=X_t'\1\{C_t = 0\} + Z_t \1\{C_t =1\}$ which the learner observes.
	The learner leverages this observation to choose another arm at the next step in order to maximize the total cumulative reward obtained after $n$ steps.
	In Algorithm~\ref{alg:cap}, we outline a pseudocode of this framework.
	\begin{algorithm}[h]
	\begin{algorithmic}[1]
		\Require{$\varepsilon \in [0,1/2)$ and $q \ge 2$}
		\State \textbf{Input:} {$P_1,\dots,P_k\in \mathcal{P}_{[q]}$ be the uncorrupted reward distributions and $H_1,\dots,H_k\in \mathcal{P}$ be the corrupted reward distributions.}
		\For{$t=1,\dots,n$}
		\State	Player plays an arm $A_t \in \{1,\dots,k\}$
		\State	Nature draws a Bernoulli $C_t \sim Ber(\varepsilon)$
		\State	Generate a corrupted reward $Z_t\sim H_{A_t}$ and an uncorrupted reward $X_t' \sim P_{A_t}$
		\State	Player observe the reward $X_t=X_t'\1\{C_t = 0\} + Z_t \1\{C_t =1\}$
		\EndFor
	\end{algorithmic}
		\caption{Bandits corrupted by Nature}\label{alg:cap}
	\end{algorithm}

	\begin{remark}[Non-adversarial corruption.]
		In the setting of \textit{Bandits corrupted by Nature}, we consider that the reward received by the learner is corrupted when $C_t=1$ and non-corrupted otherwise. Since the law of $C_t$ is a Bernoulli $Ber(\varepsilon)$, the corruption is stochastic, and independent on other variables. This is in contrast with \emph{adversarial} setups, where corruption is typically chosen by an opponent and possibly depending on other variables. Assuming a non-adversarial behavior of the Nature seem more justified than assuming an adversarial setup in applications, such as agriculture where corruption is often due to external disturbances, such as pests appearance or weather hazards, whose occurrence are typically non-adversarial. Now when corruption happens, we do not put restriction on the level of corruption. For example, we can imagine a pest outburst or hail, that may have huge impact on a crop but does not occur adversarially.
	\end{remark}

	\begin{remark}[Weak assumption on inliers] Let us highlight that we do not assume sub-Gaussian behavior for the inlier distributions $P_i$. Instead, we consider only a weak moment assumption, i.e. \emph{the inlier distributions $P_i$ have a finite variance}. Thus, our setting is capable of modeling both the heavy-tailed and corrupted settings. We highlight this generality in the regret lower bounds and empirical performance analysis in Section~\ref{sec:lower_bounds} and~\ref{sec:examples}.
	\end{remark}

	\paragraph{Corrupted regret.} In this setting, we observe that a corrupted reward distribution ($(1-\varepsilon)P_i + \varepsilon H_i$) might not have finite mean, unlike the true $P_i$'s. Thus, the regret with respect to the corrupted reward distributions might fail to quantify the goodness of the policy and its immunity to corruption while learning.
	
	In this setup, the natural notion of expected regret is measured with respect to the  mean of the non-corrupted environment $\nu$ specified by $\lbrace P_i\rbrace_{i=1}^k$. We define the regret of learning algorithm playing strategy $\pi$ after $n$ steps of interaction with the environment $\nu^\varepsilon$ as
	\begin{equation}\label{eq:def_regret}
		\kR_n(\pi,\nu^\varepsilon) \triangleq n \max_i \E_{P_i}[X'] - \E\left[ \sum_{t=1}^n X'_t\right].\tag{Corrupted regret}
	\end{equation}
	The expectation is crucially taken on  $X'_i\sim P_i$ and $X'_t\sim P_{A_t}$ but not on $X_i$ and $X_t$. The expectation on the right also incorporates possible randomization from the learner. Thus, \eqref{eq:def_regret} quantifies the loss in the rewards accumulated by policy $\pi$ from the inliers while learning only from the \textit{corrupted rewards} and also not knowing the arm with the best \textit{true reward} distribution. Thus, this definition of corrupted regret quantifies the rate of learning of a bandit algorithm as regret does for non-corrupted bandits.
	A similar notion of regret is considered in \citep{kapoor2019corruption} that deals with bounded stochastic corruptions.
	
	Due to the non-adversarial nature of the corruption, the regret can be decomposed, as in classical stochastic bandits, to make appear the expected number of pulls of suboptimal arms $\E_{\nu^\varepsilon}\left[T_i(n)\right]$, which allow us to focus the regret analysis on bounding these terms.
	\begin{Lemma}[Decomposition of corrupted regret]\label{lem:decomp_regret}
		In a corrupted environment $\nu^{\varepsilon}$, the regret writes
		\begin{align*}
			\kR_n(\pi,\nu^\varepsilon) = \sum_{i=1}^k \Delta_i \E_{\nu^\varepsilon}\left[T_i(n)\right],
		\end{align*}
		where $T_i(n)\triangleq \sum_{t=1}^n \1\{A_t=i\}$ denotes the number of pulls of arm $i$ until time $n$  and the problem-dependent quantity $\Delta_i \triangleq \max\limits_j \mu_j - \mu_i$ is called the suboptimality gap of arm $i$.
	\end{Lemma}

\section{Lower bounds for uniformly good policies under heavy-tails and corruptions}\label{sec:lower_bounds}

	In order to derive the lower bounds, it is classical to consider \emph{uniformly good} policies on some family of environments, \cite{lairobbins1985}. We introduce below the corresponding notion for corrupted environments with the set of laws $\mathfrak{D}^{\otimes k}=\mathcal{D}_1 \otimes\dots \otimes \mathcal{D}_k$, where $\mathcal{D}_i \subset  \mathcal{P}$ for each $i\in \{1,\dots, k\}$.

	\begin{definition}[Robust uniformly good policies]
		Let $\mathfrak{D}^{\otimes k}(\varepsilon)=\mathcal{D}_1(\varepsilon) \otimes\dots \otimes \mathcal{D}_k(\varepsilon)$ be a family of corrupted bandit environments on $\R$.
		For a corrupted environment $\nu^\varepsilon \in \kD^{\otimes k}(\varepsilon)$ with corresponding uncorrupted environment $\nu$, let $\mu_i(\nu)$ denote the mean reward of arm $i$ in the uncorrupted setting and $\mu_{\star}(\nu)\triangleq\max_a \mu_i(\nu)$ denote the maximum mean reward. A policy $\pi$ is uniformly good on $\mathfrak{D}^{\otimes k}(\varepsilon)$ if for any $\alpha \in (0,1]$,
		$$\forall \nu \in \mathfrak{D}^{\otimes k}(\varepsilon), \forall i \in \{1,\dots,k\}, \mu_i(\nu)<\mu_{\star}(\nu) \Rightarrow \quad \E_{\nu^\varepsilon}[T_i(n)]= o(n^\alpha). $$
	\end{definition}
	Since the corrupted setup is a special case of stochastic bandits, a lower bound can be immediately recovered with classical results, such as Lemma~\ref{lem:lower_bound_uniformly_good} below, that is a version of the change of measure argument~\citep{burnetas1997optimal}, and can be found in~\citep[Lemma 3.4]{maillard:tel-02077035}.

	\begin{Lemma}[Lower bound for uniformly good policies]\label{lem:lower_bound_uniformly_good}
		Let $\kD^{\otimes k}=\mathcal{D}_1 \otimes\dots \otimes \mathcal{D}_k$, where $\mathcal{D}_i \subset  \mathcal{P}$ for each $i\in \{1,\dots, k\}$ and let $\nu \in \kD^{\otimes k}$. Then,  any uniformly good policy on $\kD^{\otimes k}$ must pull arms such that for any $P_i \in \mathcal{D}_i$, $i \in\{1,\dots,k\}$,
		$$\forall i \in\{1,\dots,k\}, \, \mu_i \le \mu_{\star}(\nu) \quad  \Rightarrow \quad \lim\inf_{n \to \infty}\frac{\E_{\nu}[T_i(n)]}{\log(n)}\ge \frac{1}{\cK_i(P_i, \mu(P^*))}. $$
		where $\cK_i(P_i, \mu(P^*))=\inf \{ \KL(P_i, \nu): \nu_i \in \mathcal{D}_i, \mu(\nu_i)\ge \mu(P^*) \}$.
	\end{Lemma}

Lemma~\ref{lem:lower_bound_uniformly_good} is used in the traditional bandit literature to obtain lower bound on the regret using the decomposition of regret from Lemma~\ref{lem:decomp_regret}. In our setting however, the lower bound is more complex as it involves optimization on the non-convex set $\mathcal{P}_{[2]}$ of distributions with a bounded variance. It also involves an optimization in both the first and second term of the KL because we consider the worst-case corruption in both the optimal arm $P^*$ and non-optimal arm $P_i$. In this section, we do not solve these problems, but we propose lower bounds derived from the study of a specific class of heavy-tailed distributions on one hand (Lemma~\ref{lem:bound_KL_student}) and the study of a specific class of corrupted (but not heavy-tailed) distributions on the other hand (Lemma~\ref{lem:kl_cor_bernoulli}). 

Using the fact that $\cK_i(P_i, \mu(P^*))$ is an infimum that is smaller than the $\KL$ for the choice $\nu=P^*$, Lemma~\ref{lem:lower_bound_uniformly_good} induces the following weaker lower-bound:
	\begin{equation}\label{eq:lower_bound_klinf}
		\forall i \in\{1,\dots,k\}, \, \mu_i \le \mu_{\star}(\nu) \quad  \Rightarrow \quad \lim\inf_{n \to \infty}\frac{\E_{\nu}[T_i(n)]}{\log(n)}\ge \frac{1}{\KL(P_i, P^*)}.
	\end{equation}
	Equation~(\ref{eq:lower_bound_klinf}) shows that \textit{it is sufficient to have an upper bound on the $\KL$-divergence of the reward distributions interacting with the policy to get a lower bound on the number of pulls of a sub-optimal arm}.

In order to bound the $\KL$-divergence, we separately focus on two families of reward distributions, namely Student's distribution without corruption and corrupted Bernoulli distribution, that reflect the hardness due to heavy-tails and corruptions, respectively.
	\paragraph{Student's distribution without corruption.}
	To obtain a lower bound in the heavy-tailed case we use Student distributions. Student distribution are well adapted because they exhibit a finite number of finite moment which makes them heavy-tailed, and we can easily change the mean and variances of Student distribution without changing its shape parameter $d$. We denote by $\mathcal{T}_d$ the set of Student distributions with $d$ degrees of freedom,
$$\mathcal{T}_d = \left\{ P\in \mathcal{P}, \, P \text{ has distribution defined for }t \in \R\text{ by } p(t)=\frac{\Gamma(\frac{d+1}{2})}{\Gamma(d/2)\sqrt{d\pi}}\left(1+\frac{t^2}{d} \right)^{-\frac{d+1}{2}}\right\} .$$

	\begin{Lemma}[Control of KL-divergence for Heavy-tails]\label{lem:bound_KL_student}
		Let $P_1, P_2$ be two Student distributions with $d> 1$ degrees of freedom with $\E_{P_1}[X]=0$ and $\E_{P_2}[X]=\Delta$. Then,
		\begin{align}
		\KL(P_1, P_2)\le
		\begin{cases}
			\frac{3^{d-1}(d+1)^2 \Delta^2}{5\sqrt{d}}  & \text{if } \Delta \le 1\,,\\
			(d+1)\log\left(\Delta\right) + \log\left(3^{d}\frac{(d+1)^2}{5\sqrt{d}} \right) & \text{if } \Delta > 1\,.
		\end{cases}    
		\end{align}
		\end{Lemma}

	\paragraph{Corrupted Bernoulli distributions.}
	Now, we study the cost of corruption using the corrupted Bernoulli distributions. Let $P_0, P_1$ be two Bernoulli distributions on $\{0, 1\}$ such that $\P_{P_0}(1)\!=\!\P_{P_1}(0)\!>\!\P_{P_0}(0)=\P_{P_1}(1)$. We corrupt both $P_0$ and $P_1$ with a proportion $\varepsilon>0$ to get
	$Q_0 \triangleq (1-\varepsilon)P_0 + \varepsilon \delta_c$ and $Q_1\triangleq(1-\varepsilon)P_1 + \varepsilon \delta_0 $. We obtain Lemma~\ref{lem:kl_cor_bernoulli} that illustrates three bounds on $\KL(Q_0, Q_1)$ as functions of the sub-optimality gap $\Delta\triangleq \E_{P_0}[X]-\E_{P_1}[X]$, variance $\sigma^2\triangleq\Var_{P_0}(X)=\Var_{P_1}(X)$, and corruption proportion $\varepsilon$.

	\begin{Lemma}[Control of KL-divergence for Corruptions]\label{lem:kl_cor_bernoulli}
		There exists $P_0, P_1$ two Bernoulli probability distribution with $\Delta=\E_{P_0}[X]-\E_{P_1}[X]$ and $\sigma^2=\Var_{P_0}(X)=\Var_{P_1}(X)$ for which there exists $Q_0$ and $Q_1$ some $\varepsilon$-corruptions of $P_0$ and $P_1$ respectively, that have \textit{shifted sub-optimality gap} given by $\overline{\Delta}_\varepsilon=\E_{Q_0}[X]-\E_{Q_1}[X]=\Delta(1-\varepsilon) - 2\varepsilon \sigma$.
		Furthermore, they can be chosen so as to satisfy

		\noindent$\bullet$ \textbf{Uniform Bound.} For any $\Delta,\sigma$, we have
		\begin{equation}\label{eq:kl_ber_large_delta}
			\KL(Q_0, Q_1)\le (1-2\varepsilon)\log\left(1+ \frac{1-2\varepsilon}{\varepsilon}\right).
		\end{equation}

		\noindent$\bullet$ \textbf{High Distinguishability/Low Variance Regime.} If $2\sigma\frac{\varepsilon}{\sqrt{1-2\varepsilon}}<\Delta < 2\sigma$, we get		\begin{equation}\label{eq:kl_ber_small_delta}
			\KL(Q_0, Q_1)\le\frac{\overline{\Delta}_\varepsilon}{2\sigma}\log\left(1+ \frac{\overline{\Delta}_\varepsilon}{2\sigma-\overline{\Delta}_\varepsilon}\right).
		\end{equation}

		\noindent$\bullet$ \textbf{Low Distinguishability/High Variance Regime.} If $\Delta\le 2\sigma\frac{\varepsilon}{\sqrt{1-2\varepsilon}}$, there exists $\varepsilon'\le \varepsilon$ and $Q_0',Q_1'$ some $\varepsilon'$- versions of $P_0$ and $P_1$ such that $\KL(Q_0',Q_1')=0$.
	\end{Lemma}

\begin{figure}[t!]
	    \centering
	    \includegraphics{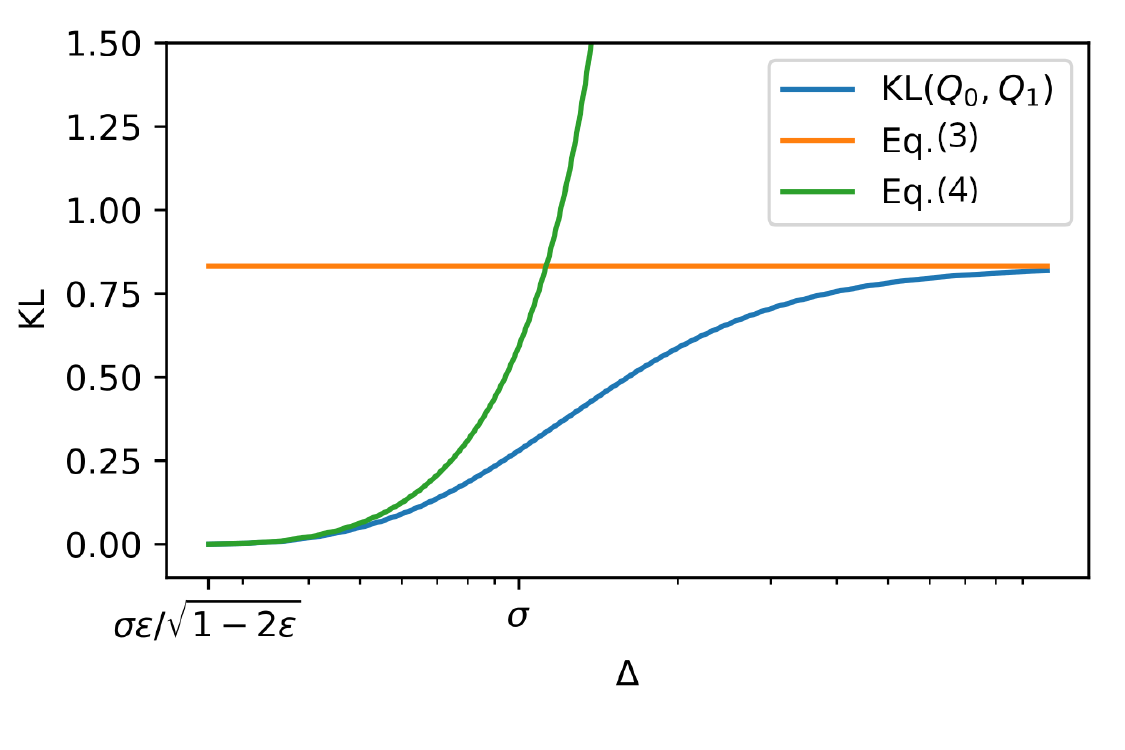}
	    \caption{Visualizing the KL and the corresponding bounds in Lemma~\ref{lem:kl_cor_bernoulli} for $\sigma = 1$ and $\varepsilon=0.2$ ($x$ axis is in log scale).}
	    \label{fig:kl_bernoulli}
	\end{figure}

	\noindent\textbf{Consequences of Lemma~\ref{lem:kl_cor_bernoulli}.} We illustrate the bounds of Lemma~\ref{lem:kl_cor_bernoulli} in {\color{red}Figure~\ref{fig:kl_bernoulli}}. The three upper bounds on the KL-divergence of corrupted Bernoullis provide us some insights regarding the impact of corruption.

	\noindent1. \textit{Three Regimes of Corruption:} We observe that depending on $\Delta/\sigma$, we can categorize the corrupted environment in three categories. For $\Delta/\sigma \in [2,+\infty)$, we observe that the KL-divergence between corrupted distributions $Q_0$ and $Q_1$ is upper bounded by a function of only corruption proportion $\varepsilon$ and is independent of the uncorrupted distributions.
	Whereas for $\Delta/\sigma \in (2\varepsilon/\sqrt{1-2\varepsilon},2)$, the distinguishability of corrupted distributions depend on the distinguishibility of uncorrupted distributions and also the corruption level. We call this the High Distinguishability/Low Variance Regime.
	For $\Delta/\sigma \in [0,2\varepsilon/\sqrt{1-2\varepsilon}]$, we observe that the KL-divergence can always go to zero. We refer to this setting as the Low Distinguishability/High Variance Regime.

	\noindent2. \textit{High Distinguishability/Low Variance Regime:} In Lemma~\ref{lem:kl_cor_bernoulli}, we observe that the effective gap to distinguish the optimal arm to the closest sub-optimal arm that dictates hardness of a bandit instance has shifted from the uncorrupted gap $\Delta$ to a\textit{ corrupted suboptimality gap}: $\overline{\Delta}_\varepsilon\triangleq \Delta(1-\varepsilon) - 2\varepsilon \sigma$.

	\noindent3. \textit{Low Distinguishability/High Variance Regime:} We notice also that there is a limit for $\Delta$ below which the corruption can make the two distributions $Q_0$ and $Q_1$ indistinguishable, this is a general phenomenon in the setting of testing in corruption neighborhoods~\citep{huber1965robust}. 



	\noindent\textbf{From KL Upper bounds to Regret Lower Bounds.} Substituting the results of Lemma~\ref{lem:bound_KL_student} and \ref{lem:kl_cor_bernoulli} in Equation~\eqref{eq:lower_bound_klinf} yield the lower bounds on regret of any uniformly good policy in heavy-tailed and corrupted settings, where reward distributions either belong to the class of corrupted student distributions or the class of corrupted Bernoulli distributions, respectively.
	We denote
	$$\mathfrak{D}_{\mathcal{T}_2}^{\otimes k} \triangleq \mathcal{T}_2 \otimes \dots \otimes \mathcal{T}_2,$$
	where $\mathcal{T}_2$ is the set of Student distributions with more than $2$ degrees of freedoms. We also define
	$$\mathfrak{D}_{\mathcal{B}(\varepsilon)}^{\otimes k} \triangleq \mathcal{B}(\varepsilon) \otimes \dots \otimes \mathcal{B}(\varepsilon), $$
	where $\mathcal{B}(\varepsilon) = \{(1-\varepsilon)P+\varepsilon H; \, H \sim Ber(p)\text{ and } P  \sim Ber(p'), p, p' \in [0,1]\}$ is the set of corrupted Bernoulli distributions.
	\begin{Theorem}[Lower bound for heavy-tailed and corrupted bandit]\label{th:lower_bound}

		Let $i$ be a suboptimal arm such that $\E_{P_i}[X]\le \max_{a}\E_{P_a}[X]$ and denote $\Delta_i\triangleq \E_{P_i}[X]- \max_{a}\E_{P_a}[X]$ and $\overline{\Delta}_{i,\varepsilon}\triangleq \Delta_i(1-\varepsilon) - 2\varepsilon \sigma_i$.

		\noindent \textbf{Student's distributions.}
		Suppose that the arms are pulled according to a policy that is uniformly good on $\mathfrak{D}_{\mathcal{T}_2}^{\otimes k}$. Then,
		\begin{equation}
			\lim\inf_{n \to \infty}\frac{\E_{\nu^\varepsilon}[T_i(n)]}{\log(n)}\ge  \frac{\sigma_i^2}{51\Delta_{i}^2} \vee \frac{1}{4\log(\Delta_i/\sigma_i)+22}.\label{eq:lower_bound_heavy}
		\end{equation}

		\noindent \textbf{Corrupted Bernoulli distributions}: Suppose that the arms are pulled according to a policy that is uniformly good on $\mathfrak{D}_{\mathcal{B}(\varepsilon)}^{\otimes k}$. Then, we have for $2\sigma_i \frac{\varepsilon}{\sqrt{1-2\varepsilon}}<\Delta_i<2\sigma_i$, then
		\begin{equation}
			\lim\inf_{n \to \infty}\frac{\E_{\nu^\varepsilon}[T_i(n)]}{\log(n)}\ge  \frac{2\sigma_i}{\overline{\Delta_{i,\varepsilon}}\log\left(1+\frac{\overline{\Delta_{i,\varepsilon}}}{2\sigma_i-\overline{\Delta_{i,\varepsilon}}}\right)},\label{eq:lower_bound_ber1}
		\end{equation}
		and for $\Delta_i>2\sigma_i$,
		\begin{equation}
		    \lim\inf_{n \to \infty}\frac{\E_{\nu^\varepsilon}[T_i(n)]}{\log(n)}\ge  \frac{1}{(1-2\varepsilon)\log\left(\frac{1-\varepsilon}{\varepsilon}\right)}.\label{eq:lower_bound_ber2}
		    \end{equation}
	\end{Theorem}
	For brevity, the detailed proof is deferred to Appendix~\ref{sec:proof_th_lower}.

	\paragraph{Small gap versus large gap regimes.}

    Due to the restriction in the family of distributions considered in Theorem~\ref{th:lower_bound}, the lower bounds are not tight and may not exhibit the correct rate of convergence for all families of distributions. However, this theorem provide some insights about the difficulties that one may encounter in corrupted and heavy-tail bandits problems, including the logarithmic dependence on $n$.

	In Theorem~\ref{th:lower_bound}, if $\Delta_i$ is small, we see that in the heavy-tailed case (Student's distribution), we recover a term very similar to the lower bound when the arms are from a Gaussian distribution. 	Now in the case where $\Delta_i$ is large, the number of suboptimal pulls in the heavy-tail setting is  $\Omega\left(1/\log\left(\frac{\Delta_i}{\sigma_i}\right)\right)$. This is the price to pay for heavy-tails.
	
	If  we are in the high distiguishability/low variance regime, i.e. $\frac{\overline{\Delta}_{i,\varepsilon}}{2\sigma_i} \in (\frac{\varepsilon}{\sqrt{1-2\varepsilon}},1)$, we recover a logarithmic lower bound which depends on a \textit{corrupted gap between means} $\overline{\Delta}_{i,\varepsilon}=\Delta_i(1-\varepsilon) -2 \varepsilon \sigma_i$. Since the corrupted gap is always smaller than the true gap $\Delta_i$, this indicates that a corrupted bandit ($\varepsilon >0$) must incur higher regret than a uncorrupted one ($\varepsilon =0$). For $\varepsilon=0$, this lower bound coincides with the lower bound for Gaussians with uncorrupted gap of means $\Delta_i$ and variance $\sigma_i^2$. On the other hand, if $\frac{\overline{\Delta}_{i,\varepsilon}}{2\sigma_i}$ is larger than $1$, we observe that we can still achieve logarithmic regret but the hardness depends on only the corruption level $\varepsilon$, specifically $\frac{1}{(1-2\varepsilon)\log\left(\frac{1-\varepsilon}{\varepsilon}\right)}$.

	\section{Robust bandit algorithm: Huber's estimator and upper bound on the regret}\label{sec:upper_bound}
	In this section, we propose an UCB-type algorithm, namely \HubUCB, addressing the Bandits corrupted by Nature setting (Algorithm~\ref{alg:huber_bandit}). This algorithm uses primarily a robust mean estimator called Huber's estimator (Section~\ref{sec:huber}) and corresponding confidence bound to develop \HubUCB{} (Section~\ref{sec:huber_confidence}). 
	We further provide a theoretical analysis in Theorem~\ref{th:upper_bound} leading to upper bound on regret of \HubUCB. We observe that the proposed upper bound matches the lower bound in Theorem~\ref{th:lower_bound} under some settings. 
	\subsection{Robust mean estimation and Huber's estimator}\label{sec:huber}
	We begin with a presentation of the Huber's estimator of mean~\citep{Huber1964RobustEO}.
	
	As we aim to design a UCB-type algorithm, the main focus is to obtain an empirical estimate of the mean
	rewards. Since the rewards are heavy-tailed and corrupted in this
	setting, we have to use a robust estimator of mean. We choose to use
	Huber's estimator~\citep{Huber1964RobustEO}, an M-estimator that is
	known for its robustness properties and have been extensively studied (e.g. the concentration properties~\citep{catoni2012challenging}).

	Huber's estimator is an M-estimator, which means that it can be derived as a minimizer of some loss function. Given access to $n$ i.i.d. random variables $X_1^n \triangleq \{X_1,\dots,X_n\}$, we define Huber's estimator as
	\begin{equation}\label{eq:huber_loss}
	    \Hub_\beta(X_1^n) \in \arg\min_{\theta \in \R} \sum_{i=1}^n \rho_\beta(X_i-\theta),
	\end{equation}
	where $\rho_\beta$ is Huber's loss function with parameter $\beta >0$. $\rho_\beta$ is a loss function that is quadratic near $0$ and linear near infinity, with $\beta$ thresholding between the quadratic and linear behaviors. 
	
	In the rest of the paper, rather than using the aforementioned definition, we represent the Huber's estimator as a root of the following equation~\citep{mathieu2021concentration}:
	\begin{equation}\label{eq:def_huber}
		\sum_{i=1}^n \psi_\beta\left(X_i-\Hub_\beta(X_1^n)\right)=0.
	\end{equation}
	Here, $\psi_\beta(x) \triangleq x \1\{|x|\le \beta\} + \beta \sign(x)\1\{|x|>\beta\}$ is called the influence function. Though the representations in Equation~\eqref{eq:huber_loss} and~\eqref{eq:def_huber} are equivalent, we prefer to use representation Equation~\eqref{eq:def_huber} as we prove the properties of Huber's estimator using those of $\psi_\beta$.

	$\beta$ plays the role of a scaling parameter. Depending on $\beta$, Huber's estimator exhibits a trade-off between the efficiency of the minimizer of the square loss, i.e. the empirical mean, and the robustness of the minimizer of the absolute loss, i.e. the empirical median.

	\subsection{Concentration of Huber's estimator in corrupted setting}\label{sec:huber_confidence}
	Let use denote the true Huber mean for a distribution $P$ as $\Hub_\beta(P)$. This means that for a random variable $Y$ with law $P$, $\Hub_\beta(P)$ satisfies $\E[\psi_\beta(Y-\Hub_\beta(P))]=0$. 	
	
	We now state our first key result on the concentration of Huber's estimator around $\Hub_\beta(P)$ in a corrupted and Heavy-tailed setting.
	\begin{Theorem}[Concentration of Empirical Huber's estimator]\label{th:concentration_huber}
		Suppose that $X_1,\dots,X_n$ are i.i.d. with law $(1-\varepsilon)P+\varepsilon H$ for some $P,H \in \mathcal{P}$ and proportion of outliers $\varepsilon\in (0,1/2)$, and $P$ has a finite variance $\sigma^2$. 
		Then,  with probability larger than $1-5\delta$,
		$$|\Hub_\beta(X_1^n)-\Hub_\beta(P)|\le \frac{ \sigma\sqrt{\frac{2 \ln(1/\delta)}{n}}+ \beta \frac{\ln(1/\delta)}{3n}+2\beta\overline{\varepsilon}\sqrt{\frac{\ln(1/\delta)}{n}}+2\beta\varepsilon}{\left(p-\sqrt{\frac{\ln(1/\delta)}{2n}}-\varepsilon\right)_+}.$$
		Here, $p = \P_P(|Y-\E_P[Y]|\le \beta/2)$ with $p> 5\varepsilon$, $\beta > 4\sigma$, $\overline{\varepsilon}=\sqrt{\frac{(1-2\varepsilon)}{\log\left( \frac{1-\varepsilon}{\varepsilon}\right)}}$, and
		$\delta \ge \exp\left(- n \frac{128\left( p-5\varepsilon \right)^2}{49\left(1+2\overline{\varepsilon}\sqrt{2}\right)^2 }\right).$
	\end{Theorem}
	
	Theorem~\ref{th:concentration_huber} gives us the concentration of $\Hub_\beta(X_1^n)$ around $\Hub_\beta(P)$, i.e. the Huber functional of the \textit{inlier} distribution $P$. This theorem will allow us to construct a UCB-type algorithm to solve the Bandits corrupted by Nature.
	
	For convenience of notation, hereafter, we denote the rate of convergence of $\Hub_\beta(X_1^n)$ to $\Hub_\beta(P)$ as
	\begin{align}\label{eq:def_rn}
	    r_n(\delta) \triangleq \frac{ \sigma\sqrt{\frac{2 \ln(1/\delta)}{n}}+ \beta \frac{\ln(1/\delta)}{3n}+2\beta\overline{\varepsilon}\sqrt{\frac{\ln(1/\delta)}{n}}+2\beta\varepsilon}{\left(p-\sqrt{\frac{\ln(1/\delta)}{2n}}-\varepsilon\right)_+} .
	\end{align}
	
	\noindent\textbf{Discussion.} Now, we provide a brief discussion on the implications of Theorem~\ref{th:concentration_huber}.
	
	\noindent \textit{1. Value of $p$:}  For most laws that exhibit concentration properties, the constant $p$ is close to $1$ as $\beta \ge 4\sigma$. One might also use Markov inequality to lower bound $p$, depending on the number of finite moments $P$ has. Bounding $p$ then becomes a trade-off on the value of $\beta$, where large values of $\beta$ implies that $p$ is close to $1$. But larger $\beta$ also leads to a less robust estimator, since the error bound in Theorem~\ref{th:concentration_huber} increases with $\beta$.
	
	\noindent\textit{2. Tightness of constants:} If there are no outliers ($\varepsilon=0$), the optimal rate of convergence in such a setting is at least of order $\sigma\sqrt{2\ln(1/\delta)/n}$ due to the central limit theorem. Theorem~\ref{th:concentration_huber} shows that we are very close to attaining this optimal constant in the leading $1/\sqrt{n}$ term. This result for Huber's estimator echoes the one presented in~\citep{catoni2012challenging}.
	
	\noindent \textit{3. Value of $\beta$:} $\beta$ is a parameter that achieve a trade-off between accuracy in the light-tailed uncorrupted setting and robustness. For our result, $\beta$ must be at least of the order of $4\sigma$. We provide a detailed discussion on the choice of $\beta$ in Section~\ref{sec:discussion}.
	
	\noindent \textit{4. Restriction on the values of $\delta$:} In Theorem~\ref{th:concentration_huber}, $\delta$ must be at least of order $e^{-n}$. This restriction may seem arbitrary but it is in fact unavoidable as shown in \citep[Theorem 4.3]{devroye2016sub}. This is a limitation of robust mean estimation that enforces our algorithm to perform a forced exploration in the beginning.
	
	\noindent \textit{5. Restriction on the values of $\varepsilon$:} In Theorem~\ref{th:concentration_huber}, $\varepsilon$ can be at most $p/5$, which implies that it is smaller than $1/5$. This restriction is common in robustness literature. In particular, in \cite{kapoor2019corruption}, $\varepsilon$ is supposed smaller than $\Delta/\sigma$. In robustness literature, \cite{lecue2020robust} and \cite{dalalyan2019outlier} assumed that $\varepsilon \le 1/768$ and $1/400$ respectively. In contrast, our analysis can handle $\varepsilon$ up to $0.2$, which is significantly higher than the existing restrictions.

	\paragraph{Bias of Huber's Estimate.} If $P$ is symmetric, we have $\Hub_\beta(P)=\E[X]$. When $P$ is non-symmetric, we need to control the distance of the Huber's estimate from the true mean, i.e. $|\Hub_\beta(P)-\E[X]|$. We call it the bias of Huber's estimate. We need to bound this bias to get a concentration of the empirical Huber's estimate $\Hub_\beta(X_1^n)$ around the true mean $\E[X]$. We control the bias using the following lemma, which is a direct consequence of~\citep[Lemma 4]{mathieu2021concentration}.
	\begin{Lemma}[Bias of Huber's estimator]\label{lem:bias_huber}
		Let $Y$ be a random variable with $\E[|Y|^q]<\infty$ for $q\ge 2$ and suppose that $\beta^2 \ge 9 \mathrm{Var}(Y)$. Then
		$$|\E[Y]-\Hub_\beta(P)|\le  \frac{2\E[|Y-\E[Y]|^q]}{(q-1)\beta^{q-1}}.$$
	\end{Lemma}
	Using Lemma~\ref{lem:bias_huber} and Theorem~\ref{th:concentration_huber}, we can control the deviations of $\Hub_\beta(X_1^n)$ from $\E[X]$. This allows us to formulate an index-based algorithm (UCB-type algorithm) for corrupted Bandits. We present this algorithm in Section~\ref{sec:algo}.

	\subsection{\HubUCB: Algorithm and regret bound}\label{sec:algo}
	In this section, we describe a robust, UCB-type algorithm called \HubUCB. We denote $\mu_i$ as the mean of arm $i$ and its variance as $\sigma_i^2$. We assume that we know the variances of the reward distributions. We refer to Section~\ref{sec:discussion} for a discussion on the choice of the parameters when the reward distributions are unknown.

	\paragraph{\HubUCB: The algorithm.}
	In order to deploy the Huber's estimator in the multi-armed bandits setting, we need to estimate the mean of the rewards of each arm separately. We do that by defining a parameter $\beta_i$ for each arm and estimating separately each $\mu_i$ using
	$$\Hub_{i,s}=\Hub_{\beta_i}\left(X_{t}, \quad 1\le t\le s \quad \text{ such that }\quad  A_t=i, \right).$$
    
	Now, at each step $t$, we define a confidence bound for arm $i$ with $s$ number of pulls as
	\begin{equation}\label{eq:ucb_bonus}
	   B_i(s,t)\triangleq \begin{cases}r_{s}(1/t^2)+b_i  & \text{if }s \ge s_{lim}(t)\\
		\infty& \text{if } s  < s_{lim}(t)
	\end{cases},
	\end{equation}	
	where $r_{s}(1/t^2)$ is defined by Equation~\eqref{eq:def_rn}, $s_{lim}(t)=\log(t)\frac{98}{128\left( p-5\varepsilon \right)^2}\left(1+2\sqrt{2}\left(\overline{\varepsilon} \vee \frac{9}{14\sqrt{2}}\right)\right)^2$, $\overline{\varepsilon}=\sqrt{\frac{(1-2\varepsilon)}{\log\left( \frac{1-\varepsilon}{\varepsilon}\right)}}$, and $b_i$ is a bound on the bias $|\E[X]-\Hub_{\beta_i}(P_i)|$. $b_i$ is zero if $P_i$ is symmetric and controlled by Lemma~\ref{lem:bias_huber} otherwise. For example, one can assign $b_i = 2\sigma_i^2/\beta_i$ by imposing $q=2$, i.e. finite second moment, in Lemma~\ref{lem:bias_huber}.

	Now, we propose \HubUCB\ that selects an arm $a_t$ at step $t$ based on the index
	\begin{equation}
	    I^{\HubUCB}_i(t)=\Hub_{i, T_i(t-1)}+B_i(T_i(t-1), t).\label{eqn:index}
	\end{equation}
	The index of \HubUCB\ together with the confidence bound defined in Equation~\eqref{eq:ucb_bonus} dictates that if an arm is less explored, i.e. $T_i(t-1) < s_{lim}(t)$, we choose that arm, and if multiple arms satisfy this, we break the tie randomly. As $t$ grows and for all the arms $T_i(t-1) \geq s_{lim}(t)$ is satisfied, we choose the arms according to the adaptive bonus. Thus, \HubUCB\ induces an initial forced exploration to obtain confident-enough robust estimates followed by a time-adaptive selection of arms.
	We present a pseudocode of \HubUCB\ in Algorithm~\ref{alg:huber_bandit}.
	\begin{algorithm}[h!]
	\caption{\HubUCB}\label{alg:huber_bandit}
	\begin{algorithmic}[1]
		\Require{$\varepsilon \in [0,1/2)$ and $\beta_i >0$, $i\le K$}
		\For{$t=1,\dots,n$}
		\State	Compute index $I^{\HubUCB}_i(t)$ (Equation \eqref{eqn:index}) for $i \in \{1,\dots, k\}$ using $X_1,\dots, X_{t-1}$.
		\State	Choose arm $a_t \in \arg\max_{i} I_i(t)$.
		\State	Observe a reward $X_t$.
		\EndFor
	\end{algorithmic}
	\end{algorithm}
	
	\noindent\textbf{Regret Analysis.} Now, we provide a regret upper bound for \HubUCB.

	\begin{Theorem}[Upper Bound on number of pulls of suboptimal arms with \HubUCB]\label{th:upper_bound}
		Suppose that for all $i$, we have $P_i\in\cP_{[2]}$, i.e. a reward distribution with finite variance $\sigma_i^2$.
		We assign $\beta_i \ge 4\sigma_i$ and $p = \inf_{1\le i\le k}\P_{P_i}(|X-\E_{P_i}[X]|\le \beta_i/2)$ such that $p > 5\varepsilon$ and $\varepsilon < 1/5$. We denote $\DeltaEpsi= (\Delta_i - 2b_i)(p-\varepsilon) - 8\beta_i \varepsilon>0$ and $\sqrt{\frac{(1-2\varepsilon)}{\log\left( \frac{1-\varepsilon}{\varepsilon}\right)}}\le \overline{\varepsilon}$.

		\noindent$\bullet$ If $\DeltaEpsi > 12\frac{\sigma_i^2}{\beta_i}\left(\sqrt{2} + 2\frac{\beta_i}{\sigma_i}\overline{\varepsilon} \right)^2$, then

		$$ \E[T_i(n)]\le \log(n)\max\!\!\left(\frac{32\beta_i}{3\DeltaEpsi}, \frac{4}{\left( p\!-\!5\varepsilon \right)^2}\left(1+2\sqrt{2}\left(\overline{\varepsilon}\vee \frac{9}{14\sqrt{2}} \right)\right)^2\right)+10(\log(n)\!+\!1)  $$

		\noindent$\bullet$ If $\DeltaEpsi \le  12\frac{\sigma_i^2}{\beta_i}\left(\sqrt{2} + 2\frac{\beta_i}{\sigma_i}\overline{\varepsilon} \right)^2$, then
		$$\E[T_i(n)]\le \log(n)\max\!\!\left( \frac{50\sigma_i^2}{9\DeltaEpsi^2}\!\left(\! \sqrt{2}\!+\!2 \frac{\beta_i}{\sigma_i}\overline{\varepsilon}\right)^{\!2}\!\!, \frac{4}{\left( p\!-\!5\varepsilon \right)^2}\!\left(\!1\!+\!2\sqrt{2}\!\left(\overline{\varepsilon}\vee\!\frac{9}{14\sqrt{2}}\!\right)\right)^{\!2} \right)+10(\log(n)\!+\!1) . $$
	\end{Theorem}
        Using Theorem~\ref{th:upper_bound} and Lemma~\ref{lem:decomp_regret}, a bound on the corrupted regret of $\HubUCB$ follows immediately.
	
	We now state a simplified version of Theorem~\ref{th:upper_bound} with worse but explicit constants for easier comprehension. Let us fix $ \beta_i^2\!=\!16\sigma_i^2$ and $\varepsilon \!\le\! 1/10$ such that $\overline{\varepsilon}= 4/(5\sqrt{\ln(9)})\simeq 0.54$, and $p \ge 1\!-\!\frac{4\sigma_i^2}{\beta_i^2}\!\ge\! \frac{3}{4}\ge 5\varepsilon \!+\! \frac{1}{4}$. Now, if we further assume that $P_i$ symmetric leading to $b_i\!=\!0$, it yields the following upper bounds.
	\begin{Corollary}[Simplified version of Theorem~\ref{th:upper_bound}]\label{cor:upper_bound}
		Suppose that for all $i$, $P_i$ is a symmetric distribution with finite variance $\sigma_i^2$. Let also denote $\DeltaEpsi \triangleq \Delta_i\left(p-\varepsilon \right)- 32 \sigma_i \varepsilon $ for $\varepsilon < 1/10$.

		\noindent$\bullet$ If $\DeltaEpsi > 6\sigma_i\left(1 + 4\sqrt{2}\overline{\varepsilon} \right)^2$, then
		\begin{align*}
			\E[T_i(n)]\le 43\log(n)\max\left(\frac{\sigma_i}{\DeltaEpsi}, 10\right)+10(\log(n)+1).
		\end{align*}
		\noindent$\bullet$ If $\DeltaEpsi  \le  6\sigma_i\left(1 + 4\sqrt{2}\overline{\varepsilon} \right)^2$, then
		\begin{align*}
			\E[T_i(n)]\le 23 \log(n)\max\left( \frac{\sigma_i^2}{\DeltaEpsi^2}\left(1+32\overline{\varepsilon}^2\right), 18\right)+10(\log(n)+1).
		\end{align*}

	\end{Corollary}

	Remark that in this corollary, we replaced some occurrences of $\overline{\varepsilon}$ by its upper bound, which is also an upper bound on $\varepsilon$. Thus, the presented result is loose up to constants but lend itself to easier comprehension. 

	\paragraph{Discussions on the Upper Bound.} Here, we discuss how this proposed upper bound of \HubUCB\ matches and mismatches with the lower bounds in Theorem~\ref{th:lower_bound}.
	
	1. \textit{Order-optimality of Upper Bound.} \HubUCB\ achieves the logarithmic regret prescribed by the lower bound (Theorem~\ref{th:lower_bound}) plus some additive error due to the fact that this is a UCB-type algorithm. Thus, \HubUCB\ is order optimal with respect to $n$.
	
	2. \textit{Two Regimes of Upper Bound.} When $\Delta_i$ is small compared to $\sigma_i$, we obtain an upper bound $\E[T_i(n)] \underset{n \rightarrow \infty}{=} \mathcal{O}\left(\log(n)\left(\frac{\sigma_i^2}{\DeltaEpsi^2} \overline{\varepsilon}^2\right)\right)$ from Corollary~\ref{cor:upper_bound}.  $\overline{\varepsilon}^2$ is of the same order of magnitude as Equation~\eqref{eq:lower_bound_ber2} because we take $\varepsilon$ strictly smaller than $1/2$. $\overline{\varepsilon}^2$ acts as an indicator of the corruption level. The term $\frac{\sigma_i^2}{\DeltaEpsi^2}$ indicates the hardness due to the corrupted gaps $\DeltaEpsi$ and echoes the hardness term $\frac{\sigma_i^2}{\Delta_i^2}$ that appears in regret upper bound of UCB for uncorrupted bandits. The hardness term $\frac{\sigma_i^2}{\DeltaEpsi^2}$ also appears in the corrupted lower bound (Equation~\eqref{eq:lower_bound_ber1}) as well as the heavy-tailed lower bound (Equation~\eqref{eq:lower_bound_heavy}) for $\Delta_i \ll \sigma_i$\footnote{We observe that the lower bound in Equation~\eqref{eq:lower_bound_heavy} depends on $\frac{\sigma_i^2}{\overline{\Delta_{i,\varepsilon}}^2}$ for $\Delta_i \ll \sigma_i$, since the first order approximation of $\log(1+x)$ is $x$ as $x \rightarrow 0$.}. 
	
	On the other hand, if $\Delta_i$ is larger than $\sigma_i$, we get that $\E[T_i(n)] = O\left(\log(n)\left(\frac{\sigma_i}{\DeltaEpsi}\vee \overline{\varepsilon}^2\vee 1\right)\right) $. This upper bound reflects the lower bound in Equation~\eqref{eq:lower_bound_ber2} that holds for $\Delta_i > 2 \sigma_i$. This reinstates the fact that for large enough suboptimality gaps, the regret of \HubUCB\ depends solely on the corruption level than the suboptimality gap. 
	
	3. \textit{Deviation from the Lower Bound.} The two regimes defined in the upper bound does not follow the exact distinctions made in the lower bounds. We observe that in upper bound, the distinction between regimes depend on a shifted suboptimality gap $\DeltaEpsi \triangleq \Delta_i\left(p-\varepsilon \right)- 32 \sigma_i \varepsilon$, while the lower bound depends on the corrupted suboptimality gap $\overline{\Delta}_{i,\varepsilon} \triangleq \Delta_i\left(1-\varepsilon \right)- 2\sigma_i \varepsilon$. This difference in constants hinder the hardness regimes and corresponding constants in upper and lower bounds to match for all $\Delta_i, \sigma_i,$ and $\varepsilon$.
	This deviation also comes from the fact that the lower bounds proposed in Theorem~\ref{th:lower_bound} consider effects of heavy-tails and corruptions separately, while the upper bound of \HubUCB\ consider them in a coupled manner.
	
	Additionally, we observe that regret of \HubUCB\ is suboptimal due to the constant additive error, which appears due to the initial forced exploration of \HubUCB\ up to $s_{lim}(t)$. Our concentration bounds and corresponding regret analysis shows that this forced exploration phase is unavoidable in order to be able to handle the case $\Delta_i \leq \sigma_i$ with \HubUCB. Removing this discrepancy between the lower and upper bounds would constitute an interesting future work.

	\subsection{Computational Details}\label{sec:discussion}
	Here, we discuss the three hyperparameters that \HubUCB\ depends on and also its computational cost.
	
	\noindent \textit{Choice of $\sigma$ and $\varepsilon$.}  In Theorem~\ref{th:upper_bound}, we assume to know the $\sigma$ and $\varepsilon$. In practice, these are unknown and we estimate $\sigma^2$ with a robust estimator of the variance, such as the median absolute deviation. In contrast, estimating $\varepsilon$ is hard. 
	There exists some heuristics, for example using proportion of point larger than 1.5 times the inter-quartile range or using more complex algorithms like Isolation Forest algorithm but these methods work in general using the hypothesis that outliers are in some way points that are located  outside of ``the bulk of the data" which conflicts with the fact that we don't suppose anything on the outliers. Moreover even
though there are heuristics, the problem of finding what constitute ``the bulk
of the data" is closely linked to problems such as finding a ``Robust minimum volume ellipsoid" which is NP-hard in general~\citep{mittal2022finding}. We refer to Appendix~\ref{sec:choice_param} for an ablation study on the choice of $\varepsilon$.

	\noindent \textit{Choice of $\beta$.} Ideally, $\beta$ should be larger than $\max_i\{4\sigma_i\}$. We recommend using the estimator of $\sigma$ to estimate a good value of $\beta$. The choice of $\beta$ reflects the difference between heavy-tailed bandits and corrupted bandits. When the data are heavy-tailed but not corrupted, \cite{catoni2012challenging} shows that $\beta\!\simeq\!\sigma\sqrt{n}$ is a good choice for the scaling parameter. However, this choice is not robust to outliers and yields a linear regret in our setup (see Section~\ref{sec:examples}). When there is corruption, $\beta$ must remains bounded even when the sample size goes to infinity in order to retain robustness. In Appendix~\ref{sec:choice_param}, we present an ablation study on the choice of $\varepsilon$.
	
	\noindent \textit{Computational Cost.} Huber's estimator has linear complexity due to the involved Iterated Re-weighting Least Squares algorithm, which is not sequential. We have to do this at every iteration, which leads \HubUCB\ to have a quadratic time complexity. This is the computational cost of using a robust mean estimator, i.e. the Huber's estimator.

\section{\SeqHubUCB: A Faster Robust Bandit Algorithm}\label{sec:sequentialhub}
In this section, we present a sequential approximation of the Huber's estimator, and we leverage it further to create a robust bandit algorithm with linear-time complexity algorithm. Here, we describe the algorithm (\SeqHubUCB) and its theoretical properties. 

\paragraph{A sequential approximation of Huber's estimator.}
The central idea is to compute the Huber's estimator using the full historical data only in logarithmic number of steps than at every step, and in between two of these re-computations, update the estimator using only the samples observed at that step.
This allows us to propose a sequential approximation of Huber's estimator, i.e. $\SHub_t$, with lower computational complexity.

By fixing the update step $P_2(t)=2^{\left\lfloor \frac{\log(t)}{\log(2)} \right\rfloor}$ before a given step $t>0$, we define the estimator $\SHub_t$ by $\SHub_0 = 0$ and
\begin{align}
\SHub_t = \begin{cases} H_t & \text{ if } t=P_2(t),\\
H_t+\frac{\sum_{i=P_2(t)}^t \psi(X_i - H_t) }{ \sum_{i=1}^t \psi'(X_i-H_t) }& \text{ otherwise.}
\end{cases}\label{eq:seqhub}    
\end{align}
Here, $H_t \triangleq \Hub(X_1^{P_2(t)})$ and $\psi$ is the influence function defined in Equation~\eqref{eq:def_huber}.
$\SHub_t$ can be conceptualized as a first order Taylor approximation of  $\Hub(X_1^t)$ around $\Hub(X_1^{P_2(t)})$. 

One might argue that $\SHub_t$ is not fully sequential rather a phased estimator as we still recompute the Huber's estimator following a geometric schedule. Thus, we still need to keep all the data in memory, leading to linear space complexity as the non-sequential Huber's estimator. 
But it features the good property of having a linear time complexity when computed using the prescribed geometric schedule. This implies that the \SeqHubUCB\ algorithm leveraging the sequential Huber's estimator achieves a linear time complexity.

\noindent\textbf{Concentration Properties of $\SHub$.} Now, in order to propose \SeqHubUCB\, we first aim to derive the rate of convergence of $\SHub_t$ towards the true Huber's mean $\Hub(P)$.
	
\begin{Theorem}\label{th:seq_huber}
If the assumptions of Theorem~\ref{th:concentration_huber} hold true, with probability larger than  $1-14\delta$, we have
\begin{align}\label{eq:conf_seqhub}
    \left|\SHub_t - \Hub(P)\right|\le  r_t(\delta)+\left(\frac{1}{p -\sqrt{\frac{\log(1/\delta)}{2t}}- \varepsilon} -1 \right) r_{P_2(t)}(\delta)
\end{align}
for any $t>0$, and $\delta\ge \exp\left(- P_2(t) \frac{128\left( p-5\varepsilon \right)^2}{49\left(1+2\overline{\varepsilon}\sqrt{2}\right)^2 }\right)$.
Here, $r_{t}(\delta)$ is defined as in Equation~\eqref{eq:def_rn}.
\end{Theorem}
We observe that the confidence bound of $\SHub_t$ includes the confidence bound of $\Hub_t$, i.e. $r_{t}(\delta)$, and an additive term proportional to $r_{P_2(t)}(\delta)$. Since $r_{P_2(t)}(\delta) \geq r_{t}(\delta)$ for $t \geq P_2(t)$, we can show that $\left|\SHub_t - \Hub(P)\right| \leq \left(p -\sqrt{\frac{\log(1/\delta)}{2t}}- \varepsilon \right)^{-1}r_{P_2(t)}(\delta)$. Thus, we obtain larger confidence bounds for $\SHub$ than that of $\Hub$, and they differ approximately by a multiplicative constant $(p-\varepsilon)^{-1}$ as $t\rightarrow \infty$.

\paragraph{\SeqHubUCB: The algorithm.}
Now, we plug-in the sequential Huber's estimator, $\SHub$, and the corresponding confidence bound (Equation~\eqref{eq:conf_seqhub}), instead of the Huber's estimator and the corresponding confidence bound in the \HubUCB\ algorithm. This allows us to construct the \SeqHubUCB\ algorithm that we present hereafter.

Specifically, we define the index of \SeqHubUCB\ as 
\begin{equation}
	    I^{\SeqHubUCB}_i(t)=\SHub_{i, T_i(t-1)}+B^{\SeqHubUCB}_i(T_i(t-1), t).\label{eqn:index_seqhub}
	\end{equation}
where
$$\SHub_{i,s}=\SHub\left(X_{t}, \quad 1\le t\le s \quad \text{ such that }\quad  A_t=i, \right),$$
and a confidence bound for arm $i$ with $s$ number of pulls is
$$B^{\SeqHubUCB}_i(s,t) \triangleq \begin{cases}r_{s}(1/t^2)+\left(\frac{1}{p -\sqrt{\frac{\log(1/\delta)}{2s}}- \varepsilon} -1 \right) r_{P_2(s)}(1/t^2) +b_i  & \text{if } P_2(s) \ge s_{lim}(t)\\
		\infty& \text{if } P_2(s)  < s_{lim}(t).
	\end{cases}$$
Here, $s_{lim}(t)$, $\overline{\varepsilon}$ and $b_i$ are same as defined for \HubUCB.

Similar to  Corollary~\ref{cor:upper_bound}, we now present a simplified regret upper bound for \SeqHubUCB. Retaining the setting of Corollary~\ref{cor:upper_bound}, we assume that $ \beta_i^2\!=\!16\sigma_i^2$, $\varepsilon \!\le\! 1/10$ implying  $\overline{\varepsilon}= 4/(5\sqrt{\ln(9)})\simeq 0.54$, $p \ge 1\!-\!\frac{4\sigma_i^2}{\beta_i^2}\!\ge\! \frac{3}{4}\ge 5\varepsilon \!+\! \frac{1}{4}$, and $P_i$ symmetric so that $b_i\!=\!0$. Further simplifying the constants yields the following regret upper bound for \SeqHubUCB.

\begin{figure}[t!]
		\includegraphics[width=\textwidth]{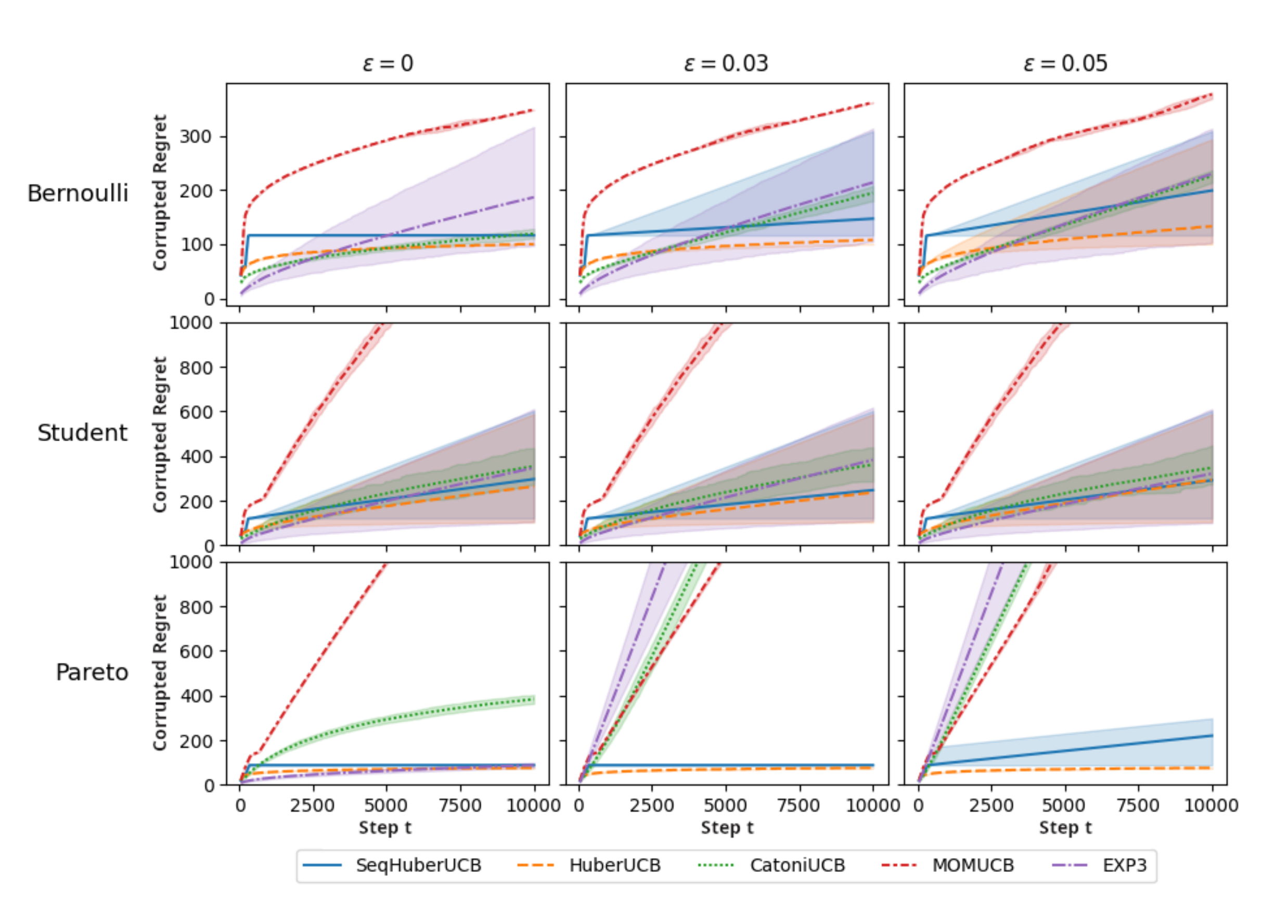}
		\caption{Cumulative regret plot of the algorithms on a corrupted Bernoulli (above), Student's (middle) and Pareto (below) reward distributions with various corruption levels $\varepsilon$. Lower corrupted regret indicates better performance for an algorithm.\label{fig:corrupted_figure}}
	\end{figure}

\begin{Lemma}[Simplified Upper Bound on Regret of \SeqHubUCB]\label{lem:upper_bound_seqhubucb}
Suppose that for all $i$, $P_i$ is a distribution with finite variance $\sigma_i^2$. Let us also denote $\DeltaEpsi =\Delta_i\left(p-\varepsilon \right)- 32 \sigma_i \varepsilon $,

\noindent$\bullet$ If $\DeltaEpsi > 18\sigma_i\left(1 + 4\sqrt{2}\overline{\varepsilon} \right)^2$, then
\begin{align*}
			\E[T_i(n)]\le 128\log(n)\max\left(\frac{\sigma_i}{\DeltaEpsi}, 2\right)+28(\log(n)+1).
		\end{align*}
		\noindent $\bullet$ If $\DeltaEpsi  \le  18\sigma_i\left(1 + 4\sqrt{2}\overline{\varepsilon} \right)^2$, then
		\begin{align*}
			\E[T_i(n)]\le 80 \log(n)\max\left( \frac{\sigma_i^2}{\DeltaEpsi^2}\left(1+32\overline{\varepsilon}^2\right), 3\right)+28(\log(n)+1).
		\end{align*}
	\end{Lemma}
	\noindent\textbf{Comparison between Regrets of \HubUCB\ and \SeqHubUCB.}
	Lemma~\ref{lem:upper_bound_seqhubucb} yields similar regret bounds for \SeqHubUCB\ as the ones obtained for \HubUCB\ in Corollary~\ref{cor:upper_bound}.
	We observe that the regrets of these two algorithms only differ in $n$-independent constants. Specifically, regret of \SeqHubUCB\ can be approximately $3-4$ times higher than that of \HubUCB. For simplicity of exposition, we present approximate constants in our results. 
	A more careful analysis might yield more fine-tuned constants. Theorem~\ref{th:seq_huber} and experimental results (Figure~\ref{fig:corrupted_figure}) indicate that it is possible to have very close performances with \SeqHubUCB\ and \HubUCB.

\section{Experimental Evaluation}\label{sec:examples}
	In this section, we assess the experimental efficiency of \HubUCB\ and \SeqHubUCB\ by plotting the empirical regret. Contrary to the uncorrupted case, we cannot really estimate the corrupted regret in \eqref{eq:def_regret} only using the observed rewards. Instead, we use the true uncorrupted gaps that we know because we are in a simulated environment, and we estimate the corrupted regret $R_n$ using
	$\sum_{i=1}^k \Delta_i \reallywidehat{T_i}(n) $,
	where $\reallywidehat{T_i}(n)=\frac{1}{M}\sum_{m=1}^M (T_i(n))_m$ is a Monte-Carlo estimation of $\E_{\nu^\varepsilon}[T_i(n)]$ over $M$ experiments.
	We use rlberry library~ \citep{Domingues_rlberry_-_A_2021} and Python3 for the experiments. We run the experiments on an 8 core Intel(R) Core(TM) i7-8665U CPU@1.90GHz. For each algorithm, we perform each experiment $100$ times to get a Monte-Carlo estimate of regret.

	\noindent\textbf{Comparison with Bandit Algorithms for Heavy-tailed and Adversarial Settings.}\label{sec:xp1}
	To the best of our knowledge, there is no existing bandit algorithm for handling unbounded stochastic corruption prior to this work. Hence, we focus on comparing ourselves to the closest settings, i.e. bandits in heavy-tailed setting and adversarial bandit algorithms. We empirically and competitively study five different algorithms: \HubUCB, \SeqHubUCB, two RobustUCB algorithms with Catoni-Huber estimator and Median of Means (MOM)~\citep{Bubeck2013BanditsWH}, and and adversarial bandit algorithm: Exp3. 
	
	\HubUCB\ is closely related to the RobustUCB with Catoni Huber estimator, which also uses Huber's estimator but with another set of parameters and confidence intervals. The RobustUCB algorithms are tuned for uncorrupted heavy-tails. Hence, they incur linear regret in a corrupted setting. This is reflected in the experiments. \textit{We also improve upon~\citep{Bubeck2013BanditsWH} as we can handle arm-dependent variances}. Exp3 is an algorithm designed for bounded Adversarial corruption, and thus, fails as the corruption is too severe.

	\noindent\textbf{Corrupted Bernoulli setting:} In Figure~\ref{fig:corrupted_figure} (above), we study a 3-armed bandits with corrupted Bernoulli distributions with means $0.1,0.97, 0.99$. The corruption applied to this bandit problem are Bernoulli distributions with means $0.999,0.999,0.001$, respectively. For \HubUCB\ and \SeqHubUCB, we choose to use $\beta_i=0.1 \sigma_i$, which seems to work better despite the theory presented before.
	We plot the mean plus/minus the standard error of the result in Figure~\ref{fig:corrupted_figure}. We do that for the three corruption proportions $\varepsilon$ equal to $0\%$, $3\%$ and $5\%$. We notice that there is a short linear regret phase at the beginning due to the forced exploration performed by the algorithms. Followed by that, \HubUCB\ and \SeqHubUCB\ incur logarithmic regret. On the other hand, Exp3, Catoni Huber Agent and MOM Agent incur logarithmic regret only in the uncorrupted setting. When the data are corrupted, i.e. $\varepsilon > 0$, their regret grow linearly.

	\noindent\textbf{Corrupted Student setting:} In Figure~\ref{fig:corrupted_figure} (middle), we study a 3-armed bandits with corrupted Student's distributions with $3$ degrees of freedom (finite second moment) and with means $0.1, 0.95, 1$. The corruption applied to this bandit problem are Gaussians with variance $1$, and means $100, 100, -1000$ respectively. For \HubUCB\ and \SeqHubUCB, we choose to use $\beta_i=\sigma_i$. The results echo the observations for the Bernoulli case except that the corruption is more drastic and affect the performance even more.

	\noindent\textbf{Corrupted Pareto setting:} In Figure~\ref{fig:corrupted_figure} (bottom), we illustrate the results for a 3-armed bandits with corrupted Pareto distributions having shape parameters $3, 3, 2.1$ (i.e. they have finite second moments), and scale parameters $0.1,0.2, 0.3$ respectively. Thus, the corresponding means are $0.15,0.3$ and $0.57$ and the standard deviations are $0.09, 0.17, 1.25$, respectively. The corruption applied to this bandit problem are Gaussians with variance $1$, and centered at $100, 100, -1000$ respectively. For \HubUCB\ and \SeqHubUCB, we choose to use $\beta=1.5\sigma_i$ and we also bound the bias $b_i$ by $\sigma_i^2/\beta_i$. The results echo the observations for the Student's distributions.
	
	Thus, we conclude that \HubUCB\ incur the lowest regret among the competing algorithms in the Bandits Corrupted by Nature setting, specially for higher corruption levels $\varepsilon$.
	Also, performances of \SeqHubUCB\ and \HubUCB\ are very close except for the Pareto distributions with high corruption level.

\section{Conclusion}\vspace*{-.5em}
	In this paper, we study the setting of Bandits corrupted by Nature that
	encompasses both the heavy-tailed rewards with bounded variance and
	unbounded corruptions in rewards.
	In this setting, we prove lower bounds on the regret that shows the
	heavy-tail bandits and corrupted bandits are strictly harder than the
	usual sub-Gaussian bandits. Specifically, in this setting, the hardness
	depends on the suboptimality gap/variance regimes. If the suboptimality
	gap is small, the hardness is dictated by
	$\sigma_i^2/\overline{\Delta}_{i,\varepsilon}^2$. Here,
	${\overline{\Delta}_{i,\varepsilon}}$ is the corrupted sub-optimality
	gap, which is smaller than the uncorrupted gap $\Delta$ and thus, harder
	to distinguish. To complement the lower bounds, we design a robust
	algorithm \HubUCB\ that uses Huber's estimator for robust mean
	estimation and a novel concentration bound on this estimator to create
	tight confidence intervals. \HubUCB\ achieves logarithmic regret that
	matches the lower bound for low suboptimality gap/high variance regime.
    We also present a sequential Huber estimator that could be of independent interest and we use it to state a linear-time robust bandit algorithm, \SeqHubUCB, that presents the same efficiency as \HubUCB.
	Unlike existing literature, we do not need any assumption on a known
	bound on corruption and a known bound on the $(1+\varepsilon)$-uncentered
	moment, which was posed as an open problem in~\citep{agrawal2021regret}.

	Since our upper and lower bounds disagree in the high gap/low variance
	regime, it will be interesting to investigate this regime further.
	From multi-armed bandits, we know that the tightest lower and upper bounds depend on the KL-divergence between optimal and suboptimal reward distributions.
	Thus, it would be imperative to study KL-divergence with corrupted distributions to better understand the Bandits corrupted by Nature problem. In this paper, we have focused on a problem-dependent regret analysis for a given $\varepsilon$. In future, it would be interesting to get some insight on how to adapt to an unknown $\varepsilon$, and to perform a problem-independent ``worst-case" analysis. Also, following the reinforcement learning literature, it will be natural to extend \HubUCB\ to contextual and linear bandit settings with corruptions and heavy-tails.
	This will facilitate its applicability to practical problems, such as choosing treatments against pests.

\clearpage
\bibliographystyle{tmlr}
\bibliography{main_tmlr}

\begin{thebibliography}{39}
\providecommand{\natexlab}[1]{#1}
\providecommand{\url}[1]{\texttt{#1}}
\expandafter\ifx\csname urlstyle\endcsname\relax
  \providecommand{\doi}[1]{doi: #1}\else
  \providecommand{\doi}{doi: \begingroup \urlstyle{rm}\Url}\fi

\bibitem[Agarwal et~al.(2019)Agarwal, Bullins, Hazan, Kakade, and
  Singh]{pmlr-v97-agarwal19c}
Naman Agarwal, Brian Bullins, Elad Hazan, Sham Kakade, and Karan Singh.
\newblock Online control with adversarial disturbances.
\newblock In Kamalika Chaudhuri and Ruslan Salakhutdinov (eds.),
  \emph{Proceedings of the 36th International Conference on Machine Learning},
  volume~97 of \emph{Proceedings of Machine Learning Research}, pp.\  111--119.
  PMLR, 09--15 Jun 2019.
\newblock URL \url{https://proceedings.mlr.press/v97/agarwal19c.html}.

\bibitem[Agrawal et~al.(2021)Agrawal, Juneja, and Koolen]{agrawal2021regret}
Shubhada Agrawal, Sandeep~K Juneja, and Wouter~M Koolen.
\newblock Regret minimization in heavy-tailed bandits.
\newblock In \emph{Conference on Learning Theory}, pp.\  26--62. PMLR, 2021.

\bibitem[Auer et~al.(2002{\natexlab{a}})Auer, Cesa-Bianchi, and
  Fischer]{auer2002finite}
Peter Auer, Nicolo Cesa-Bianchi, and Paul Fischer.
\newblock Finite-time analysis of the multiarmed bandit problem.
\newblock \emph{Machine learning}, 47\penalty0 (2-3):\penalty0 235--256,
  2002{\natexlab{a}}.

\bibitem[Auer et~al.(2002{\natexlab{b}})Auer, Cesa-Bianchi, Freund, and
  Schapire]{auer2002nonstochastic}
Peter Auer, Nicolo Cesa-Bianchi, Yoav Freund, and Robert~E Schapire.
\newblock The nonstochastic multiarmed bandit problem.
\newblock \emph{SIAM journal on computing}, 32\penalty0 (1):\penalty0 48--77,
  2002{\natexlab{b}}.

\bibitem[Bogunovic et~al.(2020)Bogunovic, Krause, and
  Scarlett]{bogunovic2020corruption}
Ilija Bogunovic, Andreas Krause, and Jonathan Scarlett.
\newblock Corruption-tolerant gaussian process bandit optimization.
\newblock In \emph{International Conference on Artificial Intelligence and
  Statistics}, pp.\  1071--1081. PMLR, 2020.

\bibitem[Bouneffouf(2021)]{bouneffouf2021corrupted}
Djallel Bouneffouf.
\newblock Corrupted contextual bandits: Online learning with corrupted context.
\newblock In \emph{ICASSP 2021-2021 IEEE International Conference on Acoustics,
  Speech and Signal Processing (ICASSP)}, pp.\  3145--3149. IEEE, 2021.

\bibitem[Bourel et~al.(2020)Bourel, Maillard, and Talebi]{bourel:hal-03000664}
Hippolyte Bourel, Odalric-Ambrym Maillard, and Mohammad~Sadegh Talebi.
\newblock {Tightening Exploration in Upper Confidence Reinforcement Learning}.
\newblock In \emph{{International Conference on Machine Learning}}, Vienna,
  Austria, July 2020.
\newblock URL \url{https://hal.archives-ouvertes.fr/hal-03000664}.

\bibitem[Bubeck et~al.(2013)Bubeck, Cesa-Bianchi, and
  Lugosi]{Bubeck2013BanditsWH}
S{\'e}bastien Bubeck, Nicolo Cesa-Bianchi, and G{\'a}bor Lugosi.
\newblock Bandits with heavy tail.
\newblock \emph{IEEE Transactions on Information Theory}, 59\penalty0
  (11):\penalty0 7711--7717, 2013.

\bibitem[Burnetas \& Katehakis(1997)Burnetas and
  Katehakis]{burnetas1997optimal}
Apostolos~N Burnetas and Michael~N Katehakis.
\newblock Optimal adaptive policies for markov decision processes.
\newblock \emph{Mathematics of Operations Research}, 22\penalty0 (1):\penalty0
  222--255, 1997.

\bibitem[Catoni(2012)]{catoni2012challenging}
Olivier Catoni.
\newblock Challenging the empirical mean and empirical variance: a deviation
  study.
\newblock In \emph{Annales de l'IHP Probabilit{\'e}s et statistiques},
  volume~48, pp.\  1148--1185, 2012.

\bibitem[Dalalyan \& Thompson(2019)Dalalyan and Thompson]{dalalyan2019outlier}
Arnak Dalalyan and Philip Thompson.
\newblock Outlier-robust estimation of a sparse linear model using l1-penalized
  huber's m-estimator.
\newblock \emph{Advances in neural information processing systems}, 32, 2019.

\bibitem[Depersin \& Lecu{\'e}(2019)Depersin and Lecu{\'e}]{depersin2019robust}
Jules Depersin and Guillaume Lecu{\'e}.
\newblock Robust subgaussian estimation of a mean vector in nearly linear time.
\newblock \emph{arXiv preprint arXiv:1906.03058}, 2019.

\bibitem[Devroye et~al.(2016)Devroye, Lerasle, Lugosi, and
  Oliveira]{devroye2016sub}
Luc Devroye, Matthieu Lerasle, Gabor Lugosi, and Roberto~I Oliveira.
\newblock Sub-gaussian mean estimators.
\newblock \emph{The Annals of Statistics}, 44\penalty0 (6):\penalty0
  2695--2725, 2016.

\bibitem[Domingues et~al.(2021)Domingues, Flet-Berliac, Leurent, Ménard,
  Shang, and Valko]{Domingues_rlberry_-_A_2021}
Omar~Darwiche Domingues, Yannis Flet-Berliac, Edouard Leurent, Pierre Ménard,
  Xuedong Shang, and Michal Valko.
\newblock {rlberry - A Reinforcement Learning Library for Research and
  Education}, 10 2021.
\newblock URL \url{https://github.com/rlberry-py/rlberry}.

\bibitem[Hajiesmaili et~al.(2020)Hajiesmaili, Talebi, Lui, Wong,
  et~al.]{hajiesmaili2020adversarial}
Mohammad Hajiesmaili, Mohammad~Sadegh Talebi, John Lui, Wing~Shing Wong, et~al.
\newblock Adversarial bandits with corruptions: Regret lower bound and
  no-regret algorithm.
\newblock \emph{Advances in Neural Information Processing Systems},
  33:\penalty0 19943--19952, 2020.

\bibitem[Huber(1964)]{Huber1964RobustEO}
Peter~J. Huber.
\newblock Robust estimation of a location parameter.
\newblock \emph{Annals of Mathematical Statistics}, 35:\penalty0 492--518,
  1964.

\bibitem[Huber(1965)]{huber1965robust}
Peter~J Huber.
\newblock A robust version of the probability ratio test.
\newblock \emph{The Annals of Mathematical Statistics}, pp.\  1753--1758, 1965.

\bibitem[Huber(2004)]{huber2004robust}
Peter~J Huber.
\newblock \emph{Robust statistics}, volume 523.
\newblock John Wiley \& Sons, 2004.

\bibitem[Kamler et~al.(2016)Kamler, Nesvorna, Stara, Erban, and
  Hubert]{kamler2016comparison}
Martin Kamler, Marta Nesvorna, Jitka Stara, Tomas Erban, and Jan Hubert.
\newblock Comparison of tau-fluvalinate, acrinathrin, and amitraz effects on
  susceptible and resistant populations of varroa destructor in a vial test.
\newblock \emph{Experimental and applied acarology}, 69\penalty0 (1):\penalty0
  1--9, 2016.

\bibitem[Kapoor et~al.(2019)Kapoor, Patel, and Kar]{kapoor2019corruption}
Sayash Kapoor, Kumar~Kshitij Patel, and Purushottam Kar.
\newblock Corruption-tolerant bandit learning.
\newblock \emph{Machine Learning}, 108\penalty0 (4):\penalty0 687--715, 2019.

\bibitem[Lai \& Robbins(1985)Lai and Robbins]{lairobbins1985}
T.L Lai and Herbert Robbins.
\newblock Asymptotically efficient adaptive allocation rules.
\newblock \emph{Advances in Applied Mathematics}, 6\penalty0 (1):\penalty0
  4--22, 1985.
\newblock ISSN 0196-8858.
\newblock \doi{https://doi.org/10.1016/0196-8858(85)90002-8}.
\newblock URL
  \url{https://www.sciencedirect.com/science/article/pii/0196885885900028}.

\bibitem[Lattimore \& Szepesv{\'a}ri(2020)Lattimore and
  Szepesv{\'a}ri]{lattimore2018bandit}
Tor Lattimore and Csaba Szepesv{\'a}ri.
\newblock \emph{Bandit algorithms}.
\newblock Cambridge University Press, 2020.

\bibitem[Lecu{\'e} \& Lerasle(2020)Lecu{\'e} and Lerasle]{lecue2020robust}
Guillaume Lecu{\'e} and Matthieu Lerasle.
\newblock Robust machine learning by median-of-means: theory and practice.
\newblock \emph{The Annals of Statistics}, 48\penalty0 (2):\penalty0 906--931,
  2020.

\bibitem[Lee et~al.(2020)Lee, Yang, Lim, and Oh]{lee2020optimal}
Kyungjae Lee, Hongjun Yang, Sungbin Lim, and Songhwai Oh.
\newblock Optimal algorithms for stochastic multi-armed bandits with heavy
  tailed rewards.
\newblock \emph{Advances in Neural Information Processing Systems},
  33:\penalty0 8452--8462, 2020.

\bibitem[Lerasle et~al.(2019)Lerasle, Szab{\'o}, Mathieu, and
  Lecu{\'e}]{lerasle2019monk}
Matthieu Lerasle, Zolt{\'a}n Szab{\'o}, Timoth{\'e}e Mathieu, and Guillaume
  Lecu{\'e}.
\newblock Monk outlier-robust mean embedding estimation by median-of-means.
\newblock In \emph{International Conference on Machine Learning}, pp.\
  3782--3793. PMLR, 2019.

\bibitem[Lykouris et~al.(2018)Lykouris, Mirrokni, and
  Paes~Leme]{lykouris2018stochastic}
Thodoris Lykouris, Vahab Mirrokni, and Renato Paes~Leme.
\newblock Stochastic bandits robust to adversarial corruptioreferences 1ns.
\newblock In \emph{Proceedings of the 50th Annual ACM SIGACT Symposium on
  Theory of Computing}, pp.\  114--122, 2018.

\bibitem[Maillard(2019)]{maillard:tel-02077035}
Odalric-Ambrym Maillard.
\newblock \emph{{Mathematics of Statistical Sequential Decision Making}}.
\newblock Habilitation {\`a} diriger des recherches, {Universit{\'e} de Lille
  Nord de France}, February 2019.
\newblock URL \url{https://hal.archives-ouvertes.fr/tel-02077035}.

\bibitem[Mathieu(2021)]{mathieu2021concentration}
Timothée Mathieu.
\newblock Concentration study of m-estimators using the influence function,
  2021.

\bibitem[Medina \& Yang(2016)Medina and Yang]{medina2016no}
Andres~Munoz Medina and Scott Yang.
\newblock No-regret algorithms for heavy-tailed linear bandits.
\newblock In \emph{International Conference on Machine Learning}, pp.\
  1642--1650. PMLR, 2016.

\bibitem[Minsker(2019)]{minsker2019distributed}
Stanislav Minsker.
\newblock Distributed statistical estimation and rates of convergence in normal
  approximation.
\newblock \emph{Electronic Journal of Statistics}, 13\penalty0 (2):\penalty0
  5213--5252, 2019.

\bibitem[Minsker \& Ndaoud(2021)Minsker and Ndaoud]{minsker2021robust}
Stanislav Minsker and Mohamed Ndaoud.
\newblock Robust and efficient mean estimation: an approach based on the
  properties of self-normalized sums.
\newblock \emph{Electronic Journal of Statistics}, 15\penalty0 (2):\penalty0
  6036--6070, 2021.

\bibitem[Mittal \& Hanasusanto(2022)Mittal and Hanasusanto]{mittal2022finding}
Areesh Mittal and Grani~A Hanasusanto.
\newblock Finding minimum volume circumscribing ellipsoids using generalized
  copositive programming.
\newblock \emph{Operations Research}, 70\penalty0 (5):\penalty0 2867--2882,
  2022.

\bibitem[Pogodin \& Lattimore(2020)Pogodin and Lattimore]{pogodin2020first}
Roman Pogodin and Tor Lattimore.
\newblock On first-order bounds, variance and gap-dependent bounds for
  adversarial bandits.
\newblock In \emph{Uncertainty in Artificial Intelligence}, pp.\  894--904.
  PMLR, 2020.

\bibitem[Prasad et~al.(2019)Prasad, Balakrishnan, and
  Ravikumar]{prasad2019unified}
Adarsh Prasad, Sivaraman Balakrishnan, and Pradeep Ravikumar.
\newblock A unified approach to robust mean estimation.
\newblock \emph{arXiv preprint arXiv:1907.00927}, 2019.

\bibitem[Prasad et~al.(2020)Prasad, Balakrishnan, and
  Ravikumar]{prasad2020robust}
Adarsh Prasad, Sivaraman Balakrishnan, and Pradeep Ravikumar.
\newblock A robust univariate mean estimator is all you need.
\newblock In \emph{International Conference on Artificial Intelligence and
  Statistics}, pp.\  4034--4044. PMLR, 2020.

\bibitem[Rinkevich(2020)]{rinkevich2020detection}
Frank~D Rinkevich.
\newblock Detection of amitraz resistance and reduced treatment efficacy in the
  varroa mite, varroa destructor, within commercial beekeeping operations.
\newblock \emph{PloS one}, 15\penalty0 (1):\penalty0 e0227264, 2020.

\bibitem[Semkiw et~al.(2013)Semkiw, Skubida, and Pohorecka]{amitrazcomparison}
Piotr Semkiw, Piotr Skubida, and Krystyna Pohorecka.
\newblock The amitraz strips efficacy in control of varroa destructor after
  many years application of amitraz in apiaries.
\newblock \emph{Journal of Apicultural Science}, 57:\penalty0 107--121, 06
  2013.
\newblock \doi{10.2478/jas-2013-0012}.

\bibitem[Shao et~al.(2018)Shao, Yu, King, and Lyu]{shao2018almost}
Han Shao, Xiaotian Yu, Irwin King, and Michael~R Lyu.
\newblock Almost optimal algorithms for linear stochastic bandits with
  heavy-tailed payoffs.
\newblock \emph{Advances in Neural Information Processing Systems}, 31, 2018.

\bibitem[Wendel(1948)]{Wendel1948NoteOT}
James~G. Wendel.
\newblock Note on the gamma function.
\newblock \emph{American Mathematical Monthly}, 55:\penalty0 563, 1948.

\end{thebibliography}

\appendix
\newpage 
\part{Appendix}
\parttoc
\newpage
\section{Proof of Theorems}
\subsection{Proof of Theorem~\ref{th:lower_bound}: Regret Lower Bound}\label{sec:proof_th_lower}
The theorem is a consequence of Lemmas~\ref{lem:lower_bound_uniformly_good}, \ref{lem:bound_KL_student} and \ref{lem:kl_cor_bernoulli}.\\
From Lemma~\ref{lem:lower_bound_uniformly_good}, we have
\begin{equation}\label{eq:inf_unif_pf_th}
 \lim\inf_{n \to \infty}\frac{\E_{\nu}[T_i(n)]}{\log(n)}\ge \frac{1}{\KL(P_0,P_1)}
\end{equation}
\textbf{Student distributions}

Let $P_0, P_1$ be student distributions with parameter $d =3$ and gap $\Delta_i$ as in Lemma~\ref{lem:bound_KL_student}.
From Lemma~\ref{lem:bound_KL_student}, we get
\begin{equation}
\KL(P_0, P_1)\le
 \begin{cases}
17\Delta_i^2  & \text{if } \Delta_i \le 1\\
4\log\left(\Delta_i\right) + \log\left(50\right) & \text{if } \Delta_i > 1
\end{cases}\end{equation}
Then, using that $\log(50)\le 17$, 
$$\KL(P_0, P_1)\le 17\Delta_i^2\wedge  4\log\left(\Delta_i\right)+17.$$
Finally, use that the variance of a student with three degrees of freedom is $\sigma_i^2 = 3$ to get that 
$$\KL(P_0, P_1)\le 51\frac{\Delta_i^2}{\sigma_i^2}\wedge  4\log\left(\frac{\Delta_i}{\sigma_i}\right)+22.$$

\noindent\textbf{Bernoulli distributions}

Let $P_0,P_1$ be as in Lemma~\ref{lem:kl_cor_bernoulli} with gap $\Delta_i$ and variance $\sigma_i$. If $2\sigma_i\frac{\varepsilon}{\sqrt{1-2\varepsilon}}< \Delta_i < 2\sigma_i$, then

\begin{equation}\label{eq:kl_student_renorm}
\KL(P_0, P_1)\le
 \frac{\overline{\Delta}_{i,\varepsilon}}{2\sigma_i}\log\left(1+\frac{\overline{\Delta}_{i,\varepsilon}}{2\sigma_i-\overline{\Delta}_{i,\varepsilon}} \right)\wedge (1-2\varepsilon)\log\left(1+\frac{1-2\varepsilon}{\varepsilon} \right) \end{equation}

Use Equation~\eqref{eq:inf_unif_pf_th} to conclude.
\subsection{Proof of Theorem~\ref{th:concentration_huber}: Concentration of Huber's Estimator}
First, we control the deviations of Huber's estimator using the deviations of $\psi_\beta(X-\Hub_\beta(X_1^n))$. We will need the following lemma to control the variance of $\psi_\beta(X-\Hub_\beta(X_1^n))$, which will in turn allow us to control its deviation with Lemma~\ref{lem:tif_tT}.

\begin{Lemma}[Controlling Variance of Influence of Huber's Estimator]\label{lem:bound_vpsi}
	Suppose that $Y_1,\dots,Y_n$ are i.i.d with law $P$. Then
	$$\Var(\psi_\beta(Y -\Hub_\beta(P))) \le \Var(Y)=\sigma^2 $$
\end{Lemma}

\begin{Lemma}[Concentrating Huber's Estimator by Concentrating the Influence]\label{lem:tif_tT}
	Suppose that $X_1,$ $\ldots,X_n$ are i.i.d with law $(1-\varepsilon)P+\varepsilon H$ for some $H \in \mathcal{P}$ and proportion of outliers $\varepsilon\in (0,1/2)$. Then, for any $\eta>0$ and $\lambda \in (0,\beta/2]$, we have
	\begin{equation*}\P(|\Hub_\beta(X_1^n)-\Hub_\beta(P)| \ge \lambda)
	\le   \P\left(\left|\frac{1}{n}\sum_{i=1}^n \psi_\beta(X_i-\Hub_\beta(P))\right|\ge\lambda \left(p-\eta-\varepsilon\right)_+\right)+2e^{-2n\eta^2}
	\end{equation*}
	where $p = \P(|Y-\E[X]|\le \beta/2)$.
\end{Lemma}
Then, using these Lemmas, we can prove the theorem.

\begin{npf}
	\item For any $\delta\in(0,1)$, with probability larger than $1-3\delta$, \label{step:bernstein}
	\begin{align}\label{eq:result_bernsteins}
	\left|\frac{1}{n}\sum_{i=1}^n \psi_\beta(X_i-\Hub_\beta(P))\right|&\le  \sigma\sqrt{\frac{2 \log(1/\delta)}{n}}+ \beta \frac{\log(1/\delta)}{2n}+2\beta\varepsilon + 2\beta\sqrt{\frac{\log(1/\delta)(1-2\varepsilon)}{n\log\left( \frac{1-\varepsilon}{\varepsilon}\right)}}.
	\end{align}
	\onepf{Write that $X_i=(1-W_i)Y_i+W_i Z_i$ where $W_1,\dots, W_n$ are i.i.d $\{0,1\}$ Bernoulli random variable with mean $\varepsilon$, $Y_1,\dots, Y_n$ are i.i.d $\sim P$ and $Z_1,\dots, Z_n$ are i.i.d with law $H$, we have
		\begin{align*}
		&~~~~\left|\frac{1}{n}\sum_{i=1}^n \psi_\beta(X_i-\Hub_\beta(P))\right|\\&=\left|\frac{1}{n}\sum_{i=1}^n \psi_\beta(Y_i-\Hub_\beta(P))+\frac{1}{n}\sum_{i=1}^n \1\{W_i = 1\} \left(\psi_\beta(Z_i-\Hub_\beta(P))-\psi_\beta(Y_i-\Hub_\beta(P)) \right)\right|\\
		&\le \left|\frac{1}{n}\sum_{i=1}^n \psi_\beta(Y_i-\Hub_\beta(P))\right|+2\beta\frac{1}{n}\sum_{i=1}^n \1\{W_i = 1\}
		\end{align*}
		
		Remark that by definition of $\Hub_\beta(P)$, it is defined as the root of the equation $\E[\psi_\beta(Y-\Hub_\beta(P))]=0$.
		From Bernstein's inequality, for any $\delta\in (0,1)$,
		$$\P\left(\left|\frac{1}{n}\sum_{i=1}^n \psi_\beta(Y_i-\Hub_\beta(P))\right|\ge \sqrt{\frac{2V_{\psi_\beta} \log(1/\delta)}{n}}+ \beta \frac{\log(1/\delta)}{3n}\right)\le 2\delta$$
		where $V_{\psi_\beta} = \Var(\psi_\beta(Y_i-\Hub_\beta(P)))$.

		Then, using that Bernoulli random variables with mean $\varepsilon$ are sub-Gaussian with variance parameter $\frac{1-2\varepsilon}{2\log((1-\varepsilon)/\varepsilon)}$ (see \cite[Lemma 6]{bourel:hal-03000664}),
		$$\P\left(\frac{1}{n}\sum_{i=1}^n \1\{W_i = 1\}\le \varepsilon + \sqrt{\frac{\log(1/\delta)(1-2\varepsilon)}{n\log\left( \frac{1-\varepsilon}{\varepsilon}\right)}}  \right)\ge 1-\delta.$$
		Then, using Lemma~\ref{lem:bound_vpsi} we get for any $\delta\in(0,1)$, with probability larger than $1-3\delta$,
		\begin{align}\label{eq:concentration_if}
		\left|\frac{1}{n}\sum_{i=1}^n \psi_\beta(X_i-\Hub_\beta(P))\right|&\le  \sigma\sqrt{\frac{2 \log(1/\delta)}{n}}+ \beta \frac{\log(1/\delta)}{2n}+2\beta\varepsilon + 2\beta\sqrt{\frac{\log(1/\delta)(1-2\varepsilon)}{n\log\left( \frac{1-\varepsilon}{\varepsilon}\right)}}.
		\end{align}}
	\item{Using $\eta= \sqrt{\frac{\log(1/\delta)}{2n}}$, the hypotheses of Lemma~\ref{lem:tif_tT} are verified.}\\
	\onepf{
		To apply Lemma~\ref{lem:tif_tT}, it is sufficient that
		\begin{equation}\label{eq:condition_lem8}
		\sigma\sqrt{\frac{2t}{n}}+ \beta \frac{\log(1/\delta)}{3n}+2\beta\varepsilon+2\beta\sqrt{\frac{\log(1/\delta)(1-2\varepsilon)}{n\log\left( \frac{1-\varepsilon}{\varepsilon}\right)}}\le \frac{\beta}{2}\left(p-\sqrt{\frac{\log(1/\delta)}{2n}}-\varepsilon\right)
		\end{equation}
		and using that $4\sigma \le \beta$, we have that it is sufficient that
		
		\begin{equation}\label{eq:cond1}
		\sqrt{\frac{\log(1/\delta)}{2n}}+ \frac{\log(1/\delta)}{3n}+ 2\sqrt{\frac{\log(1/\delta)(1-2\varepsilon)}{n\log\left( \frac{1-\varepsilon}{\varepsilon}\right)}}\le \frac{1}{2}\left(p-5\varepsilon\right).
		\end{equation}
		This is a polynomial in $\sqrt{\log(1/\delta)/n}$ that we need to solve. We use the following elementary algebra lemma.
		\begin{Lemma}[2nd order polynomial root bound]\label{lem:polynomial_root_bound}
			let $a,b,c$ be three positive constants and $x$ verify $ax^2+bx-c \le 0$. Suppose that $\frac{4ac}{b^2} \le d$, then $x$ must verify
			$$x \ge \frac{2c(\sqrt{d+1}-1)}{db}.$$
		\end{Lemma}
		Observe that we have
		$$\frac{2\left( p-5\varepsilon \right)}{3\left( \frac{1}{\sqrt{2}}+\frac{2\sqrt{1-2\varepsilon}}{\sqrt{\log\left( \frac{1-\varepsilon}{\varepsilon}\right)}}\right)^2}\le \frac{4}{3}$$
		and $(\sqrt{4/3+1}-1)/(4/3) \ge 8/7$, hence, from Lemma~\ref{lem:polynomial_root_bound}, we get the following sufficient condition for Equation~\eqref{eq:cond1} to hold:
		\begin{align*}
		\sqrt{\log(1/\delta)/n}\le\frac{8\sqrt{2}\left( p-5\varepsilon \right)}{7\left(1+\frac{2\sqrt{2(1-2\varepsilon)}}{\sqrt{\log\left( \frac{1-\varepsilon}{\varepsilon}\right)}}\right)}.
		\end{align*}
		Hence, taking this to the square,
		$$\log(1/\delta) \le n\frac{128\left( p-5\varepsilon \right)^2}{49\left(1+\frac{2\sqrt{2(1-2\varepsilon)}}{\sqrt{\log\left( \frac{1-\varepsilon}{\varepsilon}\right)}}\right)^2}.$$
	}
	\item Using Lemma~\ref{lem:tif_tT} and \ref{step:bernstein} prove that the theorem is true.
	\onepf{
		The hypotheses of Lemma~\ref{lem:tif_tT} are verified and we can use its result and together with Equation~\eqref{eq:result_bernsteins} we get with probability larger than $1-5\delta$,
		$$|\Hub_\beta(X_1^n)-\Hub_\beta(P)|\le \frac{ \sigma\sqrt{\frac{2 \log(1/\delta)}{n}}+ \beta \frac{\log(1/\delta)}{3n}+2\beta\sqrt{\frac{\log(1/\delta)(1-2\varepsilon)}{n\log\left( \frac{1-\varepsilon}{\varepsilon}\right)}}+2\beta\varepsilon}{\left(p-\sqrt{\frac{\log(1/\delta)}{2n}}-\varepsilon\right)_+}.$$}
\end{npf}

\subsection{Proof of Theorem~\ref{th:seq_huber}: Concentration of Sequential Huber's Estimator}\label{sec:proof_seq_huber}
In this proof, we denote 
$$r_t(\delta):=\frac{ \sigma\sqrt{\frac{2 \log(1/\delta)}{t}}+ \beta \frac{\log(1/\delta)}{3t}+2\beta\overline{\varepsilon}\sqrt{\frac{\log(1/\delta)}{t}}+2\beta\varepsilon}{\left(p-\sqrt{\frac{\log(1/\delta)}{2t}}-\varepsilon\right)_+}$$
this is the rate of convergence of $\Hub_\beta(X_1^t)$ to $\Hub_\beta(P)$, as stated by Theorem~\ref{th:concentration_huber}.

Let $P_2(t)<t<P_2(t+1)$, define 
$$f_t(u)=\frac{1}{t}\sum_{i=1}^t \psi_\beta(X_i-u).$$
$f_t$ is a continuous function, we take its derivative in distribution to get that
\begin{align*}
f_t(\Hub_\beta(P))=f_t(H_t)+(\Hub_\beta(P)-H_t)f_t'(H_t)+\int_{H_t}^{\Hub_\beta(P)} f_t''\left(u\right)(\Hub_\beta(P)-u)\d u 
\end{align*}
Then, by definition of $\SHub_t$, we also have
$$0=f_t(H_t)+(\SHub_t-H_t)f_t'(H_t).$$
Hence, 
\begin{equation}\label{eq:error_seqhub}
f_t(Hub(P)) = ( \Hub_\beta(P)-\SHub_t)f_t'(H_t)+\int_{H_t}^{\Hub_\beta(P)} f_t''\left(u\right)(\Hub_\beta(P)-u)\d u .
\end{equation}
where $f_t'(u)=-\frac{1}{t}\sum_{i=1}^t \1\{|X_i-u|\le \beta\}$ and $f_t''(u) = -\frac{1}{t}\sum_{i=1}^t (\delta_{X_i-u-\beta}-\delta_{X_i-u+\beta})$ where $\delta_x$ is the dirac mass in $x$.

$f_t'(H_t)$ is a sum of indicator functions and should be blose to $\P(|X-\E[X]|\le \beta)$, which is close to $1$. 

\textbf{Bound on $f_t'(H_t)$}

We bound $| f_t'(H_t)|$. We have 
\begin{align*}
| f_t'(H_t)|&= \frac{1}{t}\sum_{i=1}^t \1\{|X_i-H_t|\le \beta\}\\
&\ge \frac{1}{t}\sum_{i=1}^t \1\{|X_i-\Hub_\beta(P)|\le \beta-|H_t-\Hub_\beta(P)|\}.
\end{align*}

Choose the limiting $\delta$ which is $\delta=\exp\left(- P_2(t) \frac{128\left( p-5\varepsilon \right)^2}{49\left(1+2\overline{\varepsilon}\sqrt{2}\right)^2 }\right)$, from Equation~\eqref{eq:condition_lem8}, we get that  $r_t(\delta)\le \beta/2$.

Then, we have from Theorem~\ref{th:concentration_huber},  with probability larger than
$1-5\exp\left(- P_2(t) \frac{128\left( p-5\varepsilon \right)^2}{49\left(1+2\overline{\varepsilon}\sqrt{2}\right)^2 }\right)$, that $|H_t - \Hub_\beta(P)|\le \beta/2$, and then,
\begin{align}\label{eq:fpHtfpHub}
| f_t'(H_t)|\ge \frac{1}{t}\sum_{i=1}^t \1\{|X_i-\Hub_\beta(P)|\le \beta/2\}.
\end{align}


\textbf{Bound on the integral of $f_t''$}.

We have,
\begin{align*}
\int_{H_t}^{\Hub_\beta(P)}& f_t''\left(u\right)(\Hub_\beta(P)-u)\d u \\
&=   \frac{1}{t}\sum_{i=1}^t \int_{H_t}^{\Hub_\beta(P)}\left(\delta_{X_i-u-\beta}-\delta_{X_i-u+\beta}\right)(\Hub_\beta(P)-u)\d u\\
&=\frac{1}{t}\sum_{i=1}^t (\Hub_\beta(P)-X_i-\beta)\1\{X_i\in I_- \} - (\Hub_\beta(P)-X_i+\beta)\1\{X_i\in I_+\} 
\end{align*}
where $I_-$ and $I_+$ are the two undirected intervals 
$$I_- = [H_t -\beta, \Hub_\beta(P)-\beta]\quad \text{and}\quad I_+=[H_t +\beta, \Hub_\beta(P)+\beta].$$
\begin{figure}[h!]
\begin{center}
\begin{tikzpicture}
\draw[very thick,-{latex}] (0,0) -- (14,0) node[below]{};

  \draw (1,3pt) -- (1,-3pt) node[above,yshift=3pt]{$\Hub_\beta(P)-\beta$};
 \draw (3,3pt) -- (3,-3pt) node[above,yshift=3pt]{$H_t-\beta$};

  \draw (6,3pt) -- (6,-3pt) node[above,yshift=3pt]{$\Hub(P)$};
 \draw (8,3pt) -- (8,-3pt) node[above,yshift=3pt]{$H_t$};
 
   \draw (11,3pt) -- (11,-3pt) node[above,yshift=3pt]{$\Hub_\beta(P)+\beta$};
 \draw (13,3pt) -- (13,-3pt) node[above,yshift=3pt]{$H_t+\beta$};

\draw [decorate,decoration={brace,amplitude=5pt,mirror,raise=2ex}]
  (1,0) -- (3,0) node[midway,yshift=-2em]{$I_-$};

\draw [decorate,decoration={brace,amplitude=5pt,mirror,raise=2ex}]
  (11,0) -- (13,0) node[midway,yshift=-2em]{$I_+$};

\end{tikzpicture}
\caption{Illustration $I_-$ and $I_+$}
\end{center}
\end{figure}
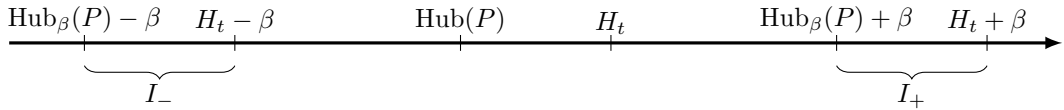

 Having that $|\Hub_\beta(X_1^t)-H_t|\le \beta/2$, we have that $I_- \cap I_+ = \emptyset$. Then, choosing either the sum $\frac{1}{t}\sum_{i=1}^t (\Hub_\beta(P)-X_i-\beta)\1\{X_i\in I_- \}$ or $\frac{1}{t}\sum_{i=1}^t (\Hub_\beta(P)-X_i-\beta)\1\{X_i\in I_+ \}$ according to which one is larger.
If $X_i \in I_+$, we have $|\Hub_\beta(P)-X_i+\beta|\le |\Hub_\beta(P)-H_t|$ and if $X_i \in I_-$, $|\Hub_\beta(P)-X_i-\beta|\le |\Hub_\beta(P)-H_t|$, hence we have
\begin{align*} &\left|\frac{1}{t}\sum_{i=1}^t\int_{H_t}^{\Hub_\beta(P)}\left(\delta_{X_i-u-\beta}-\delta_{X_i-u+\beta}\right)(\Hub_\beta(P)-u)\d u\right|\\
&\le  |\Hub_\beta(P)-H_t| \max\left( \frac{1}{t}\sum_{i=1}^t \1\{X_i \in I_-\}, \frac{1}{t}\sum_{i=1}^t \1\{X_i \in I_+ \}\right).
\end{align*}
Now remark that by Equation~\eqref{eq:fpHtfpHub}, we have 
$$\sum_{i=1}^t \1\{X_i -H_t\}=| f_t'(H_t)|\ge \frac{1}{t}\sum_{i=1}^t \1\{|X_i-\Hub_\beta(P)|\le \beta/2\}$$
Let us denote $p_t(\beta) = \frac{1}{t}\sum_{i=1}^t \1\{|X_i-\Hub_\beta(P)|\le \beta/2\}$. 

There cannot be more than $1-p_t(\beta)$ fraction of the $X_i$'s that are outside of $[H_t-\beta, H_t+\beta]$. Similarly, there cannot be more than $1-p_t(\beta)$ fraction of the $X_i's$ that are outside of $[\Hub_\beta(P)-\beta/2, \Hub_\beta(P)+\beta/2]$. Hence, if $H_t \le \Hub_\beta(P)$, then $I_- \subset [H_t-\beta, H_t+\beta]^c$ and the proportion of $X_i$'s in $I_-$ can't be larger than $1-p_t(\beta)$.

If $\Hub_\beta(P)\le H_t$, then $I_- \subset [\Hub_\beta(P)-\beta, \Hub_\beta(P)+\beta]^c$ which is itself a subset of $[\Hub_\beta(P)-\beta/2, \Hub_\beta(P)+\beta/2]^c$ and the proportion of $X_i$'s included in $[\Hub_\beta(P)-\beta/2, \Hub_\beta(P)+\beta/2]^c$ cannot be larger than $3/10$. 

In both cases, $\frac{1}{t}\sum_{i=1}^t \1\{X_i \in I_-\} \le 1-p_t(\beta)$. A similar reasoning holds for $I_+$, hence 
$$\left|\int_{H_t}^{\Hub_\beta(P)} f_t''\left(u\right)(\Hub_\beta(P)-u)\d u\right| \le(1-p_t(\beta)) |\Hub_\beta(P)-H_t|$$
Then, using Equation~\eqref{eq:fpHtfpHub} and Equation~\eqref{eq:error_seqhub}, we get with probability larger than $1-5\exp\left(- P_2(t) \frac{128\left( p-5\varepsilon \right)^2}{49\left(1+2\overline{\varepsilon}\sqrt{2}\right)^2 }\right)$,
\begin{align}\label{eq:seqhub2}
|\Hub_\beta(P)-\SHub_t| &\le \frac{f_t(\Hub_\beta(P))+\left|\int_{H_t}^{\Hub_\beta(P)} f_t''\left(u\right)(\Hub_\beta(P)-u)\d u\right|}{ f_t'(H_t)}\nonumber \\
&\le  \frac{f_t(\Hub_\beta(P))+(1-p_t(\beta))|\Hub_\beta(P)-H_t|}{p_t(\beta)}.
\end{align}
Then let $\delta \le \exp\left(- P_2(t) \frac{128\left( p-5\varepsilon \right)^2}{49\left(1+2\overline{\varepsilon}\sqrt{2}\right)^2 }\right)$, we use Equation~\eqref{eq:concentration_if} to say that with probability larger than $1-3\delta$, we have 
$$f_t(\Hub_\beta(P)) \le  \sigma\sqrt{\frac{2 \log(1/\delta)}{t}}+ \beta \frac{\log(1/\delta)}{2t}+2\beta\varepsilon + 2\beta\sqrt{\frac{\log(1/\delta)(1-2\varepsilon)}{t\log\left( \frac{1-\varepsilon}{\varepsilon}\right)}}.$$
Then, using Hoeffding's inequality after taking out the outliers, we get with probability larger than $1-\delta$, that 
$$p_t(\beta)=\frac{1}{t}\sum_{i=1}^t \1\{|X_i-\Hub(P)|\le \beta/2\}\ge p -\sqrt{\frac{\log(1/\delta)}{2t}}- \varepsilon$$
to recover that the first term of the right-hand-side of Equation~\eqref{eq:seqhub2} is smaller than $r_t(\delta)$. 
Then, using Theorem~\ref{th:concentration_huber}, we get that with probability larger than $1-5\exp\left(- P_2(t) \frac{128\left( p-5\varepsilon \right)^2}{49\left(1+2\overline{\varepsilon}\sqrt{2}\right)^2 }\right)-9\delta\ge 1-14\delta$, 
\begin{equation*}
|\Hub_\beta(P)-\SHub_t|\le r_t(\delta)+\left(\frac{1}{p -\sqrt{\frac{\log(1/\delta)}{2t}}- \varepsilon} -1 \right) r_{P_2(t)}(\delta).
\end{equation*}

\subsection{Proof of  Theorem~\ref{th:upper_bound}: Regret Upper bound of HuberUCB}\label{sec:proof_upper_bound}
If $A_t=i$ then at least one of the following four inequalities is true:
\begin{equation}\label{eq:event1}
\reallywidehat{\Hub}_{1,T_1(t-1)} + B_1( T_1(t-1), t)\le \mu_1
\end{equation}
or
\begin{equation}\label{eq:event2}
\reallywidehat{\Hub}_{i,T_i(t-1)} \ge \mu_i+B_i(T_i(t-1), t)
\end{equation}
or
\begin{equation}\label{eq:event3}
 \Delta_i <  2B_i( T_i(t-1), t)
\end{equation}
or
\begin{equation}\label{eq:event4}
T_1(t-1)< s_{lim}(t) = \frac{98\log(t)}{128\left( p-5\varepsilon \right)^2}\left(1+2\sqrt{2}\left(\overline{\varepsilon} \vee \frac{9}{14\sqrt{2}} \right) \right)^2
\end{equation}
Indeed, if $T_i(t-1)<s_{lim}(t) $, then $B_i(T_i(t-1), t)=\infty$ and Inequality~\eqref{eq:event3} is true. On the other hand, if $T_i(t-1)\ge s_{lim}(t) $, then we have $B_i(T_i(t-1),t)$ is finite and all four inequalities are false, then,
\begin{align*}
 \reallywidehat{\Hub}_{1,T_1(t-1)} + B_1( T_1(t-1), t)&> \mu_1\\
 &=\mu_i + \Delta_i\\
 &\ge \mu_i +2B_i(T_i(t-1), n)\\
  &\ge \mu_i +2B_i(T_i(t-1), t)\\
 &\ge \reallywidehat{\Hub}_{i,T_i(t-1)}  +B_i(T_i(t-1), t)\\
 \end{align*}
which implies that $A_t \neq i$.

\begin{npf}
 \item We have that $\P\left(\text{\eqref{eq:event1} is true}\right)\le 5/t$.\\
 \onepf{

Then, we have that,
\begin{align*}
 \P\left( \reallywidehat{\Hub}_{1,T_1(t-1)} + B_1( T_1(t-1), t)\le \mu_1\right)&\le \sum_{s=1}^{t} \P\left( \reallywidehat{\Hub}_{1,s} + B_1(s, t)\le \mu_1\right)\\
 &= \sum_{s=\lceil s_{lim}(t)\rceil }^{t} \P\left( \reallywidehat{\Hub}_{1,s} -\mu_1\le - B_1(s, t)\right)\\
 \end{align*}
Then, use Theorem~\ref{th:concentration_huber}, we get
 \begin{align*}
  \P\left( \reallywidehat{\Hub}_{1,T_1(t-1)} + B_1( T_1(t-1), t)\le \mu_1\right)&\le  \sum_{s=\lceil s_{lim}(t)\rceil}^t 5e^{-\log(t^2)}\\
&\le  \sum_{s=\lceil s_{lim}(t)\rceil}^t \frac{5}{t^2}\le \frac{5}{t}.
\end{align*}
 }
 \item  Similarly, for arm $i$, we have
$$\P\left( \reallywidehat{\Hub}_{i,T_i(t-1)} \ge \mu_i+B_i(T_i(t-1),t)\right) \le   \frac{5}{t}$$
\onepf{
We have,
\begin{align*}
 \P\left( \reallywidehat{\Hub}_{i,T_i(t-1)} \ge \mu_i+B_i(T_i(t-1),t)\right) &\le   \sum_{s=\lceil s_{lim}(t)\rceil }^{t} \P\left( \reallywidehat{\Hub}_{i,s} -\mu_i\ge  B_i(s, t)\right)\\
 &\le \sum_{s=\lceil s_{lim}(t)\rceil}^t 5e^{-\log(t^2)}\le \frac{5}{t}.
\end{align*}
}
\item Let $v\in \N$. If one of the two following conditions are true, then for all $t$ such that $T_i(t-1)\ge v$, we have $\Delta_i \ge   2B_i( T_i(t-1), t)$ (i.e. Equation~\eqref{eq:event3} is false).\\
Condition 1: if $\DeltaEpsi > 12\frac{\sigma_i^2}{\beta_i}\left(\sqrt{2} + 2\frac{\beta_i}{\sigma_i}\overline{\varepsilon} \right)^2$ and $v \le \log(n)\frac{96\beta_i}{9\DeltaEpsi}$.\\
Condition 2: if $\DeltaEpsi \le  12\frac{\sigma_i^2}{\beta_i}\left(\sqrt{2} + 2\frac{\beta_i}{\sigma_i}\overline{\varepsilon} \right)^2$ and
$v \le \frac{50}{9\DeltaEpsi^2}\left( \sigma_i\sqrt{2}+2\beta_i\overline{\varepsilon}\right)^2\log(n).$\\
\onepf{We search for the smallest value $v\ge s_{lim}(t)$ such that $\Delta_i$ verifies
$$ \Delta_i \ge  2B_i(v, t)=2\frac{ \sigma_i \sqrt{\frac{2 \log(t^2)}{v}}+ \beta \frac{\log(t^2)}{3v}+2\overline{\varepsilon}\beta_i\sqrt{\frac{\log(t^2)}{v}}+2\beta_i\varepsilon}{\left(p-\sqrt{\frac{\log(t^2)}{2v}}-\varepsilon\right)}+2b_i.$$

First, we simplify the expression, having that $v \ge s_{lim}(t)$, we have
$$\frac{\log(t^2)}{2v}\le \frac{128(p-5\varepsilon)^2}{98(1+9/7)^2}\le  \frac{(p-\varepsilon)^2}{4},$$
hence we simplify to
$$ \Delta_i \ge  \frac{4}{\left(p-\varepsilon\right)}\left( \sigma_i \sqrt{\frac{2 \log(t^2)}{v}}+ \beta_i \frac{\log(t^2)}{3v}+2\beta_i\overline{\varepsilon}\sqrt{\frac{\log(t^2)}{v}}+2\beta_i\varepsilon\right)+2b_i$$
let us denote $\DeltaEpsi= (\Delta_i - 2b_i)(p-\varepsilon) - 8\beta_i \varepsilon$, we are searching for $v$ such that
$$\beta_i \frac{\log(t^2)}{3v}+ \sqrt{\frac{\log(t^2)}{v}}\left( \sigma_i\sqrt{2}+2\beta_i\overline{\varepsilon}\right)-\frac{\DeltaEpsi}{4}\le 0$$
This is a second order polynomial in $\sqrt{\log(t^2)/v}$.

If $\DeltaEpsi  > 0$, then the smallest $v>0$ is

\begin{align*}
 \sqrt{\frac{\log(t^2)}{v}} &= \frac{3}{2\beta_i}\left(-\left( \sigma_i\sqrt{2}+2\overline{\varepsilon}\beta_i\right) + \sqrt{\left( \sigma_i\sqrt{2}+2\beta_i\overline{\varepsilon}\right)^2  + \frac{\DeltaEpsi \beta_i}{3} }\right).
\end{align*}

\textbf{First setting:} if $\DeltaEpsi > 12\frac{\sigma_i^2}{\beta_i}\left(\sqrt{2} + 2\frac{\beta_i}{\sigma_i}\overline{\varepsilon} \right)^2$,

In that case, we have

\begin{align*}
 \sqrt{\frac{\log(t^2)}{v}} &\ge \frac{3}{2\beta_i}\left(-\left( \sigma_i\sqrt{2}+2\beta_i\overline{\varepsilon}\right) +\sqrt{\frac{\beta_i\DeltaEpsi}{3}} \right)\ge \frac{3}{2\beta_i} \sqrt{\frac{\beta_i\DeltaEpsi}{12}}=\sqrt{\frac{9\DeltaEpsi}{48\beta_i}}
\end{align*}
Hence, $v \le \log(t)\frac{96\beta_i}{9\DeltaEpsi}$.

\textbf{Second setting:} if $\DeltaEpsi \le  12\frac{\sigma_i^2}{\beta_i}\left(\sqrt{2} + 2\frac{\beta_i}{\sigma_i}\overline{\varepsilon} \right)^2$,
then we use Lemma~\ref{lem:polynomial_root_bound}, using that
$$\frac{\DeltaEpsi \beta_i}{3\left( \sigma_i\sqrt{2}+2\beta_i\overline{\varepsilon}\right)^2} \le 4 $$
and the fact that $\frac{\sqrt{1+4}-1}{4} \ge\frac{3}{10}$, we get,
\begin{align*}
 \sqrt{\frac{\log(t^2)}{v}} &\ge \frac{3\DeltaEpsi}{5\left( \sigma_i\sqrt{2}+2\beta_i\overline{\varepsilon}\right)}
\end{align*}
Hence,
$$v \le \frac{50}{9\DeltaEpsi^2}\left( \sigma_i\sqrt{2}+2\beta_i\overline{\varepsilon}\right)^2\log(t). $$}
\item Using All the previous steps, we prove the theorem.
\onepf{We have
\begin{align*}
 \E[T_i(t)] &= \E\left[\sum_{t=1}^t \1\{A_t = i\} \right]\\
 &\le \lfloor \max(v, s_{lim}(t))\rfloor+\E\left[\sum_{t=\lfloor \max(v, s_{lim}(t)) \rfloor+1}^t\1\{A_t = i \text{ and \eqref{eq:event3} is false} \}\right]\\
 &\le \lfloor \max(v, s_{lim}(t))\rfloor+\E\left[\sum_{t=\lfloor \max(v, s_{lim}(t)) \rfloor+1}^t\1\{\text{\eqref{eq:event1} or \eqref{eq:event2} or \eqref{eq:event4} is true} \}\right]\\
  &= \lfloor \max(v, s_{lim}(t))\rfloor+\sum_{t=\lfloor \min(v, s_{lim}(t)) \rfloor+1}^t \P\left(\text{\eqref{eq:event1} or \eqref{eq:event2} is true} \right)\\
 &\le \lfloor \max(v, s_{lim}(t))\rfloor+ 2\sum_{t=\lfloor \min(v, s_{lim}(t)) \rfloor+1}^t \frac{5}{t}
\end{align*}
using the harmonic series bound by $\log(t)+1$, we have
$$ \E[T_i(t)]\le \max(v, s_{lim}(t))+10(\log(t)+1)  $$

Then, we replace the value of $v$,

\textbf{First setting:} $\DeltaEpsi > 12\frac{\sigma_i^2}{\beta_i}\left(\sqrt{2} + 2\frac{\beta_i}{\sigma_i}\overline{\varepsilon} \right)^2$

$$ \E[T_i(t)]\le \log(t)\max\left(\frac{96\beta_i}{9\DeltaEpsi}, \frac{4}{\left( p-5\varepsilon \right)^2}\left(1+2\sqrt{2}\left(\overline{\varepsilon}\vee \frac{9}{14\sqrt{2}} \right)\right)^2\right)+10(\log(t)+1)  $$

 \textbf{Second setting:} if $\DeltaEpsi \le  12\frac{\sigma_i^2}{\beta_i}\left(\sqrt{2} + 2\frac{\beta_i}{\sigma_i}\overline{\varepsilon} \right)^2$, then
$$\E[T_i(t)]\le \log(n)\max\left( \frac{50}{9\DeltaEpsi^2}\left( \sigma_i\sqrt{2}+2\beta_i\overline{\varepsilon}\right)^2, \frac{4}{\left( p-5\varepsilon \right)^2}\left(1+2\sqrt{2}\left(\overline{\varepsilon}\vee \frac{9}{14\sqrt{2}} \right)\right)^2 \right)+10(\log(t)+1) . $$
}

This concludes the proof of Theorem~\ref{th:upper_bound}.\end{npf}

\clearpage
\section{Proof of Technical Lemmas and Corollaries}

\subsection{Preliminary lemmas}
\subsubsection{Proof of Lemma~\ref{lem:decomp_regret}: Regret Decomposition}
From Equation~\eqref{eq:def_regret}, we have
$$
R_n =\sum_{a=1}^k \sum_{t=1}^n \E\left[( \max_a \E_{P_a}[X'] - X'_t)\1\left\{A_t = a \right\} \right]
$$
Then, we condition on $A_t$
\begin{align*}
\E\left[( \max_a \E_{P_a}[X'] - X'_t)\1\left\{A_t = a \right\} | A_t\right] &= \1\{A_t = a\} \E[\max_a \E_{P_a}[X']  - X'_t|A_t]\\
&= \1\{A_t = a\}(\max_a \E_{P_a}[X'] -\mu_{A_t})\\
&= \1\{A_t = a\}(\max_a \E_{P_a}[X'] -\mu_{a})= \1\{A_t = a\}\Delta_a
\end{align*}
and this stays true whatever the policy, because the policy at time $t$ use knowledge up to time $t-1$, hence its decision does not depend on $X_t$. Hence, we have
$$R_n(\pi) = \sum_{a=1}^k \Delta_a \E_{\pi(\cdot|X_1 ^n,A_1^n)}\left[T_a(n)\right] $$
where $T_a(n)$ is with respect to the randomness of $\pi$, which is to say that we compute $\E[T_i(n)]$ in the corrupted setting and not in the uncorrupted one.
$$R_n = \sum_{a=1}^k \Delta_a \E_{\nu_\varepsilon}\left[T_a(n)\right].$$

\subsubsection{Proof of Lemma~\ref{lem:bound_KL_student}: KL for Student's Distribution}\label{sec:proof_lem_KL_student}
First, we compute the $\chi^2$ divergence between the two laws $f_a$ and $f_0$. We have, for any $a \ge 0$
\begin{align*}
\mathrm{d}_{\chi^2}(f_a, f_0)&=\int \frac{(f_a(x)-f_0(x))^2}{f_0(x)}\mathrm{d}x\\
&=\frac{\Gamma\left(\frac{d+1}{2} \right)}{\Gamma\left( \frac{d}{2}\right)\sqrt{d\pi} }\int_\R \left(\frac{1}{\left(1+\frac{(x-a)^2}{d}\right)^{\frac{d+1}{2}}}-\frac{1}{\left(1+\frac{x^2}{d}\right)^{\frac{+1}{2}}}\right)^2\left(1+\frac{x^2}{d}\right)^{\frac{d+1}{2}} \mathrm{d}x\\
&=\frac{\Gamma\left(\frac{d+1}{2} \right)}{\Gamma\left( \frac{d}{2}\right)\sqrt{d\pi} }
\int_\R \frac{\left(\left(1+\frac{(x-a)^2}{d}\right)^{\frac{d+1}{2}}-\left(1+\frac{x^2}{d}\right)^{\frac{d+1}{2}}\right)^2}{\left(1+\frac{(x-a)^2}{d}\right)^{d+1}\left(1+\frac{x^2}{d}\right)^{\frac{d+1}{2}}}\mathrm{d}x\\
&=\frac{\Gamma\left(\frac{d+1}{2} \right)}{\Gamma\left( \frac{d}{2}\right)\sqrt{d\pi} }
\left(\int_\R \frac{\mathrm{d}x}{\left(1+\frac{x^2}{d}\right)^{\frac{d+1}{2}}}-2\int_\R \frac{\mathrm{d}x}{\left(1+\frac{(x-a)^2}{d}\right)^{\frac{d+1}{2}}}+\int_\R \frac{\left(1+\frac{x^2}{d}\right)^{\frac{d+1}{2}}}{\left(1+\frac{(x-a)^2}{d}\right)^{d+1}}\mathrm{d}x\right).
\end{align*}
The first two terms are respectively equal to $1$ and $-2$ using the fact that the student distribution integrate to $1$. Then, we do the change of variable $y = x-a$ in the last integral to get
\begin{align*}
\mathrm{d}_{\chi^2}(f_a, f_0)&=\frac{\Gamma\left(\frac{d+1}{2} \right)}{\Gamma\left( \frac{d}{2}\right)\sqrt{d\pi} }
\int_\R \frac{\left(1+\frac{(y+a)^2}{d}\right)^{\frac{d+1}{2}}}{\left(1+\frac{y^2}{d}\right)^{d+1}}\mathrm{d}y - 1.
\end{align*}
this is a polynomial of degree $d$ in the variable $a$. We have the following Lemma proven in Section~\ref{sec:proof_lem_integral_student}.
\begin{Lemma}\label{lem:integral_student}
For $a \ge 0$ and $d\ge0$, we have the following algebraic inequality.
	$$  \int_\R \frac{\left(1+\frac{(y+a)^2}{d}\right)^{\frac{d+1}{2}}}{\left(1+\frac{y^2}{d}\right)^{d+1}}\mathrm{d}y \le \frac{a^2}{2\sqrt{d}}(d+1)^2   \left(2+ \frac{a}{\sqrt{d}} \right)^{d-1}+\int_\R \frac{(1+y^2/d)^{\frac{d+1}{2}}}{\left(1+\frac{y^2}{d}\right)^{d+1}} \mathrm{d}y.
$$
\end{Lemma}
Using this lemma, and because we recognize up to a constant the integral of the student distribution on $\R$ in the right hand side, we have
\begin{align*}
\mathrm{d}_{\chi^2}(f_a, f_0)&=\frac{\Gamma\left(\frac{d+1}{2} \right)}{\Gamma\left( \frac{d}{2}\right)\sqrt{d\pi} }
\left(  \frac{a^2}{2\sqrt{d}}(d+1)^2   \left(2 +\frac{a}{\sqrt{d}} \right)^{d-1}+\int_\R \frac{(1+y^2/d)^{\frac{d+1}{2}}}{\left(1+\frac{y^2}{d}\right)^{d+1}} \mathrm{d}y \right) - 1\\
&\le  \frac{\Gamma\left(\frac{d+1}{2} \right)}{\Gamma\left( \frac{d}{2}\right)\sqrt{d\pi} }
\frac{a^2}{2\sqrt{d}}(d+1)^2   \left(2+ \frac{a}{\sqrt{d}} \right)^{d-1}
\end{align*}
then, use that for any $d \ge 1$, $\Gamma(\frac{d+1}{2})\le \Gamma(\frac{d}{2})\sqrt{d/2} $ from \cite{Wendel1948NoteOT}, hence
\begin{align*}
\mathrm{d}_{\chi^2}(f_a, f_0)&\le  \frac{a^2 (d+1)^2 }{2\sqrt{2d\pi} }
\left(2+ \frac{a}{\sqrt{d}} \right)^{d-1} \le \frac{a^2 (d+1)^2 }{5\sqrt{d}}
\left(2+ \frac{a}{\sqrt{d}} \right)^{d-1},
\end{align*}
using $2\sqrt{2\pi}\ge 5$·
Then, we use the link between KL divergence and $\chi^2$ divergence to get the result.
\begin{align}\label{eq:KL_student}
\KL(f_a, f_0) &\le \log(1+\mathrm{d}_{\chi^2}(f_a, f_0))\nonumber\\
&\le \log\left(1+\frac{a^2 (d+1)^2 }{5\sqrt{d}}
\left(2+ \frac{a}{\sqrt{d}} \right)^{d-1}\right)
\end{align}

Then, we have
$$\log\left(1+\frac{a^2 (d+1)^2 }{5\sqrt{d}}
\left(2+ \frac{a}{\sqrt{d}} \right)^{d-1}\right) \le
\begin{cases}
\log\left(1+3^{d-1} \frac{(d+1)^2}{5\sqrt{d}} a^2\right) & \text{if }a<1\\
\log\left(1+\frac{(d+1)^2}{5\sqrt{d}}a^{d+1}\left(\frac{(d+1)^2}{\sqrt{d}}+\frac{1}{\sqrt{d}}\right)^{d-1}\right) & \text{if }a\ge 1
\end{cases}
$$
hence, using that $1 \le 3^{d-1}\frac{(d+1)^2}{d}a^{d+1}$
$$\log\left(1+\frac{a^2 (d+1)^2 }{d}
\left(2+ \frac{a}{\sqrt{d}} \right)^{d-1}\right)  \le
\begin{cases}
3^{d-1} \frac{(d+1)^2}{5\sqrt{d}} a^2 & \text{if }a< 1\\
(d+1)\log\left(a\right) + \log\left(3^{d}\frac{(d+1)^2}{5\sqrt{d}} \right) & \text{if }a \ge 1
\end{cases}
$$
Inject this in Equation~\eqref{eq:KL_student} to get the result.

\subsubsection{Proof of Lemma~\ref{lem:kl_cor_bernoulli}: KL for Corrupted Bernoulli Distribution}
Let $\alpha \in (0,1/2)$.
Define\\
$P_0 = (1-\alpha)\delta_0 + \alpha \delta_1 $,\\
$P_1 = \alpha \delta_0 + (1-\alpha)\delta_1$,\\
$Q_0=(1-\varepsilon)(1-\alpha) \delta_0 + (1-(1-\varepsilon)(1-\alpha)) \delta_1$,\\
$Q_1 = (1-(1-\varepsilon)(1-\alpha)) \delta_0+(1-\varepsilon)(1-\alpha) \delta_1$.

One can check that
$Q_0  =(1-\varepsilon)P_0+ \varepsilon \delta_1$ and $Q_1 = (1-\varepsilon)P_1 + \varepsilon \delta_0$ and hence $Q_0$ and $Q_1$ are in the $\varepsilon$-corrupted neighborhood of respectively $P_0$ and $P_1$.

We have
\begin{align*}
\KL(Q_0, Q_1) &= \sum_{k \in \{0,c\}}\P_{Q_0}\left(X = k\right) \log\left(\frac{\P_{Q_0}\left(X = k\right)}{\P_{Q_1}\left(X = k\right)} \right)\\
&= (1-\varepsilon)(1-\alpha)\log\left( \frac{(1-\varepsilon)(1-\alpha)}{1-(1-\varepsilon)(1-\alpha)}\right)+ \left(1-(1-\varepsilon)(1-\alpha)\right)\log\left( \frac{1-(1-\varepsilon)(1-\alpha)}{(1-\varepsilon)(1-\alpha)}\right)\\
&= \left((1-\varepsilon)(1-\alpha) - \left( 1- (1-\varepsilon)(1-\alpha)\right) \right)\log\left( \frac{(1-\varepsilon)(1-\alpha)}{1-(1-\varepsilon)(1-\alpha)}\right)\\
&=\left(1-2\varepsilon-2\alpha+2\varepsilon\alpha \right)\log\left(1+ \frac{1-2\varepsilon-2\alpha+2\varepsilon\alpha}{\varepsilon+\alpha - \varepsilon\alpha}\right)
\end{align*}
Then, note that $\Delta = \E_{P_1}[X]-\E_{P_0}[X]=(1-2\alpha)$ and $\sigma^2 = \Var_{P_0}(X)=\Var_{P_1}(X)=\alpha(1-\alpha)$. Hence, with $\alpha = \frac{1}{2}\left(1-\Delta\right)$.
\begin{align}\label{eq:kl_bernoulli_cor}
\KL(Q_0, Q_1) &=\left(1-2\varepsilon-\left(1-\Delta\right)(1-\varepsilon) \right)\log\left(1+ \frac{1-2\varepsilon-\left(1-\Delta\right)(1-\varepsilon)}{\varepsilon+\frac{1}{2}\left(1-\Delta\right)(1- \varepsilon)}\right)\\
&=\left(\Delta(1-\varepsilon)-\varepsilon \right)\log\left(1+ \frac{\Delta(1-\varepsilon)-\varepsilon}{\frac{1}{2}(1+\varepsilon)-\frac{1}{2}\Delta(1- \varepsilon)}\right)
\end{align}

\textbf{Uniform bound}: if $\varepsilon>0$, we have
\begin{align*}
\KL(Q_0, Q_1) &\le \left(1-2\varepsilon \right)\log\left(1+ \frac{1-2\varepsilon}{\varepsilon}\right).
\end{align*}

\textbf{High distinguishibility regime}: in the setting $2\sigma > \Delta$, we have the bound
\begin{align*}
\KL(Q_0, Q_1) &\le \left(\frac{\Delta}{2\sigma}(1-\varepsilon)-\varepsilon \right)\log\left(1+ 2\frac{\frac{\Delta}{2\sigma}(1-\varepsilon)-\varepsilon}{1-\left(\frac{\Delta}{2\sigma}(1- \varepsilon)-\varepsilon\right)}\right)\\
&= \left(\frac{\Delta(1-\varepsilon)-2\sigma\varepsilon}{2\sigma}\right)\log\left(1+ 2\frac{\Delta(1-\varepsilon)-2\sigma \varepsilon}{2\sigma-\left(\Delta(1- \varepsilon)-2\sigma\varepsilon\right)}\right)
\end{align*}

\textbf{Low distinguishibility regime}: if $ \Delta \le  2\sigma\frac{\varepsilon}{\sqrt{1-2\varepsilon}}$. Then there exists $\varepsilon' \le \varepsilon$ such that $ \Delta = 2\sigma\frac{\varepsilon'}{\sqrt{1-2\varepsilon'}}$ and then, from Equation~\eqref{eq:kl_bernoulli_cor}, there exists $Q_0',Q_1'$ which are $\varepsilon'$-corrupted versions of $P_0$ and $P_1$ such that $KL(Q_0',Q_1')=0$

\newpage

\subsection{Lemmas for Regret upper bound}
\subsubsection{Proof of Corollary~\ref{cor:upper_bound}: Simplified Upper Bound of HuberUCB}

Replacing $\beta_i$ by $4\sigma_i$, we have

$\bullet$ If $\DeltaEpsi > 6\sigma_i\left(1 + 4\sqrt{2}\overline{\varepsilon} \right)^2$, then

$$ \E[T_i(n)]\le \log(n)\max\left(\frac{128\sigma_i}{3\DeltaEpsi}, \frac{4}{\left( p-5\varepsilon \right)^2}\left(1+2\sqrt{2}\left(\overline{\varepsilon}\vee \frac{9}{14\sqrt{2}} \right)\right)^2\right)+10(\log(n)+1)  $$

$\bullet$ If $\DeltaEpsi > 6\sigma_i\left(1 + 4\sqrt{2}\overline{\varepsilon} \right)^2$, then
$$\E[T_i(n)]\le \log(n)\max\left( \frac{50\sigma_i^2}{9\DeltaEpsi^2}\left( \sqrt{2}+8\overline{\varepsilon}\right)^2, \frac{4}{\left( p-5\varepsilon \right)^2}\left(1+2\sqrt{2}\left(\overline{\varepsilon}\vee \frac{9}{14\sqrt{2}} \right)\right)^2 \right)+10(\log(n)+1) . $$

Then, we use that
\begin{align*}
\left(1+2\sqrt{2}\left(\overline{\varepsilon}\vee \frac{9}{14\sqrt{2}} \right)\right)^2&\le 2\left(1+\left(2\sqrt{2}\left(\overline{\varepsilon}\vee \frac{9}{14\sqrt{2}} \right)\right)^2\right)\\
&=2+8\left(\overline{\varepsilon}^2 \vee \frac{81}{392}\right)\le 8\overline{\varepsilon}^2 + 2+\frac{648}{392}\le 8\overline{\varepsilon}^2+4
\end{align*}
and that $p-5\varepsilon \ge 1/4$, to get

$\bullet$ If $\DeltaEpsi > 6\sigma_i\left(1 + 4\sqrt{2}\overline{\varepsilon} \right)^2$, then

\begin{align*}
\E[T_i(n)]&\le \log(n)\max\left(\frac{128\sigma_i}{3\DeltaEpsi}, 512\overline{\varepsilon}^2+256\right)+10(\log(n)+1)\\
&= \frac{128}{3}\log(n)\max\left(\frac{\sigma_i}{\DeltaEpsi}, 12\overline{\varepsilon}^2+6\right)+10(\log(n)+1)\\
&\le 43\log(n)\max\left(\frac{\sigma_i}{\DeltaEpsi}, 12\overline{\varepsilon}^2+6\right)+10(\log(n)+1)
\end{align*}

$\bullet$ If $\DeltaEpsi > 6\sigma_i\left(1 + 4\sqrt{2}\overline{\varepsilon} \right)^2$, then
\begin{align*}
\E[T_i(n)]&\le \log(n)\max\left( \frac{50\sigma_i^2}{9\DeltaEpsi^2}\left( \sqrt{2}+8\overline{\varepsilon}\right)^2, 512\overline{\varepsilon}^2+256 \right)+10(\log(n)+1) \\
&\le \log(n)\max\left( \frac{100\sigma_i^2}{9\DeltaEpsi^2}\left(2+64\overline{\varepsilon}^2\right), 512\overline{\varepsilon}^2+256 \right)+10(\log(n)+1)\\
&\le 23 \log(n)\max\left( \frac{\sigma_i^2}{\DeltaEpsi^2}\left(1+32\overline{\varepsilon}^2\right), 24\overline{\varepsilon}^2+12 \right)+10(\log(n)+1)
\end{align*}

\subsubsection{Proof of Lemma~\ref{lem:upper_bound_seqhubucb}:  Regret Upper bound for SeqHuberUCB}

In this section we virtually copy the proof of the regret for HuberUCB done in Section~\ref{sec:proof_upper_bound} with modified constants and using the crude bound $P_2(s)\ge s/2$ whenever necessary. 

If $A_t=i$ then at least one of the following four inequalities is true:
\begin{equation}\label{eq:event1_2}
\reallywidehat{\SHub}_{1,T_1(t-1)} + B_1( T_1(t-1), t)\le \mu_1
\end{equation}
or
\begin{equation}\label{eq:event2_2}
\reallywidehat{\SHub}_{i,T_i(t-1)} \ge \mu_i+B_i(T_i(t-1), t)
\end{equation}
or
\begin{equation}\label{eq:event3_2}
 \Delta_i <  2B_i( T_i(t-1), t)
\end{equation}
or
\begin{equation}\label{eq:event4_2}
P_2(T_1(t-1))< s_{lim}(t) = \frac{98\log(t)}{128\left( p-5\varepsilon \right)^2}\left(1+2\sqrt{2}\left(\overline{\varepsilon} \vee \frac{9}{14\sqrt{2}} \right) \right)^2
\end{equation}
Indeed, if $P_2(T_i(t-1))<s_{lim}(t) $, then $B_i(T_i(t-1), t)=\infty$ and Inequality~\eqref{eq:event3_2} is true. On the other hand, if $P_2(T_i(t-1))\ge s_{lim}(t) $, then we have $B_i(T_i(t-1),t)$ is finite and all four inequalities are false, then,
\begin{align*}
 \reallywidehat{\SHub}_{1,T_1(t-1)} + B_1( T_1(t-1), t)&> \mu_1\\
 &=\mu_i + \Delta_i\\
 &\ge \mu_i +2B_i(T_i(t-1), n)\\
  &\ge \mu_i +2B_i(T_i(t-1), t)\\
 &\ge \reallywidehat{\SHub}_{i,T_i(t-1)}  +B_i(T_i(t-1), t)\\
 \end{align*}
which implies that $A_t \neq i$.

\begin{npf}
 \item We have that $\P\left(\text{\eqref{eq:event1_2} is true}\right)\le 14/t$.\\
 \onepf{

Then, we have that,
\begin{align*}
 \P\left( \reallywidehat{\SHub}_{1,T_1(t-1)} + B_1( T_1(t-1), t)\le \mu_1\right)&\le \sum_{s=1}^{t} \P\left( \reallywidehat{\SHub}_{1,s} + B_1(s, t)\le \mu_1\right)\\
 &= \sum_{s=\lceil s_{lim}(t)\rceil }^{t} \P\left( \reallywidehat{\SHub}_{1,s} -\mu_1\le - B_1(s, t)\right)\\
 \end{align*}
Then, use Theorem~\ref{th:seq_huber}, we get
 \begin{align*}
  \P\left( \reallywidehat{\SHub}_{1,T_1(t-1)} + B_1( T_1(t-1), t)\le \mu_1\right)&\le  \sum_{s=\lceil s_{lim}(t)\rceil}^t 14e^{-\log(t^2)}\\
&\le  \sum_{s=\lceil s_{lim}(t)\rceil}^t \frac{14}{t^2}\le \frac{14}{t}.
\end{align*}
 }
 \item  Similarly, for arm $i$, we have
$$\P\left( \reallywidehat{\SHub}_{i,T_i(t-1)} \ge \mu_i+B_i(T_i(t-1),t)\right) \le   \frac{14}{t}$$
\onepf{
We have,
\begin{align*}
 \P\left( \reallywidehat{\SHub}_{i,T_i(t-1)} \ge \mu_i+B_i(T_i(t-1),t)\right) &\le   \sum_{s=\lceil s_{lim}(t)\rceil }^{t} \P\left( \reallywidehat{\SHub}_{i,s} -\mu_i\ge  B_i(s, t)\right)\\
 &\le \sum_{s=\lceil s_{lim}(t)\rceil}^t 14e^{-\log(t^2)}\le \frac{14}{t}.
\end{align*}
}
\item Let $v\in \N$. If one of the two following conditions are true, then for all $t$ such that $P_2(T_i(t-1))\ge v$, we have $\Delta_i \ge   2B_i( T_i(t-1), t)$ (i.e. Equation~\eqref{eq:event3_2} is false).\\
Condition 1: if $\DeltaEpsi > 12\frac{\sigma_i^2}{\beta_i}\left(\sqrt{2} + 2\frac{\beta_i}{\sigma_i}\overline{\varepsilon} \right)^2$ and $v \le \log(t)\frac{96\beta_i}{9\DeltaEpsi}$.\\
Condition 2: if $\DeltaEpsi \le  12\frac{\sigma_i^2}{\beta_i}\left(\sqrt{2} + 2\frac{\beta_i}{\sigma_i}\overline{\varepsilon} \right)^2$ and
$v \le \frac{50}{9\DeltaEpsi^2}\left( \sigma_i\sqrt{2}+2\beta_i\overline{\varepsilon}\right)^2\log(t).$\\
\onepf{We search for the smallest value $v\ge s_{lim}(t)$ such that $\Delta_i$ verifies
$$ \Delta_i \ge  2B_i(v, t)=2r_v(1/t^2)+2\left(\frac{1}{p-\sqrt{\frac{\log(t^2)}{2v}}-\varepsilon}-1\right)r_{P_2(v)}(1/t^2)+2b_i.$$

First, we simplify the expression, having that $v \ge s_{lim}(t)$, we have
$$\frac{\log(t^2)}{2v}\le \frac{128(p-5\varepsilon)^2}{98(1+9/7)^2}\le  \frac{(p-\varepsilon)^2}{4},$$
hence $r_v(1/t^2)\le \frac{2}{\left(p-\varepsilon\right)}\left( \sigma_i \sqrt{\frac{2 \log(t^2)}{v}}\right)$ and we simplify the condition to
\begin{align*}
 \Delta_i \ge&  \frac{4}{\left(p-\varepsilon\right)}\left( \sigma_i \sqrt{\frac{2 \log(t^2)}{v}}+ \beta_i \frac{\log(t^2)}{3v}+2\beta_i\overline{\varepsilon}\sqrt{\frac{\log(t^2)}{v}}+2\beta_i\varepsilon\right) \\
&+ \frac{4}{p-\varepsilon}\left( \sigma_i \sqrt{\frac{2 \log(t^2)}{{P_2(v)}}}+ \beta_i \frac{\log(t^2)}{3{P_2(v)}}+2\beta_i\overline{\varepsilon}\sqrt{\frac{\log(t^2)}{P_2(v)}}+2\beta_i\varepsilon\right)+2b_i\\
\ge & \frac{12}{\left(p-\varepsilon\right)}\left( \sigma_i \sqrt{\frac{2 \log(t^2)}{v}}+ \beta_i \frac{\log(t^2)}{3v}+2\beta_i\overline{\varepsilon}\sqrt{\frac{\log(t^2)}{v}}+2\beta_i\varepsilon\right)+2b_i
\end{align*}
where we used that $P_2(v)\ge v/2$. 

Let us denote $\DeltaEpsi= (\Delta_i - 2b_i)(p-\varepsilon) - 24\beta_i \varepsilon$, we are searching for $v$ such that
$$\beta_i \frac{\log(t^2)}{3v}+ \sqrt{\frac{\log(t^2)}{v}}\left( \sigma_i\sqrt{2}+2\beta_i\overline{\varepsilon}\right)-\frac{\DeltaEpsi}{12}\le 0$$
This is a second order polynomial in $\sqrt{\log(t^2)/v}$.

If $\DeltaEpsi  > 0$, then the smallest $v>0$ is

\begin{align*}
 \sqrt{\frac{\log(t^2)}{v}} &= \frac{3}{2\beta_i}\left(-\left( \sigma_i\sqrt{2}+2\overline{\varepsilon}\beta_i\right) + \sqrt{\left( \sigma_i\sqrt{2}+2\beta_i\overline{\varepsilon}\right)^2  + \frac{\DeltaEpsi \beta_i}{9} }\right).
\end{align*}

\textbf{First setting:} if $\DeltaEpsi > 36\frac{\sigma_i^2}{\beta_i}\left(\sqrt{2} + 2\frac{\beta_i}{\sigma_i}\overline{\varepsilon} \right)^2$,

In that case, we have

\begin{align*}
 \sqrt{\frac{\log(t^2)}{v}} &\ge \frac{3}{2\beta_i}\left(-\left( \sigma_i\sqrt{2}+2\beta_i\overline{\varepsilon}\right) +\sqrt{\frac{\beta_i\DeltaEpsi}{9}} \right)\ge \frac{3}{2\beta_i} \sqrt{\frac{\beta_i\DeltaEpsi}{36}}=\sqrt{\frac{\DeltaEpsi}{16\beta_i}}
\end{align*}
Hence, $v \le \log(t)\frac{32\beta_i}{\DeltaEpsi}$.

\textbf{Second setting:} if $\DeltaEpsi \le  36\frac{\sigma_i^2}{\beta_i}\left(\sqrt{2} + 2\frac{\beta_i}{\sigma_i}\overline{\varepsilon} \right)^2$,
then we use Lemma~\ref{lem:polynomial_root_bound}, using that
$$\frac{\DeltaEpsi \beta_i}{9\left( \sigma_i\sqrt{2}+2\beta_i\overline{\varepsilon}\right)^2} \le 4 $$
and the fact that $\frac{\sqrt{1+4}-1}{4} \ge\frac{3}{10}$, we get,
\begin{align*}
 \sqrt{\frac{\log(t^2)}{v}} &\ge \frac{\DeltaEpsi}{20\left( \sigma_i\sqrt{2}+2\beta_i\overline{\varepsilon}\right)}
\end{align*}
Hence,
$$v \le \frac{40}{\DeltaEpsi^2}\left( \sigma_i\sqrt{2}+2\beta_i\overline{\varepsilon}\right)^2 \log(t). $$
}
\item Using All the previous steps, we prove the theorem.\\
\onepf{We have
\begin{align*}
 \E[T_i(t)] &= \E\left[\sum_{t=1}^t \1\{A_t = i\} \right]\\
 &\le \lfloor \max(v, 2s_{lim}(t))\rfloor+\E\left[\sum_{t=\lfloor \max(v, 2s_{lim}(t)) \rfloor+1}^t\1\{A_t = i \text{ and \eqref{eq:event3_2} is false} \}\right]\\
 &\le \lfloor \max(v, 2s_{lim}(t))\rfloor+\E\left[\sum_{t=\lfloor \max(v,2 s_{lim}(t)) \rfloor+1}^t\1\{\text{\eqref{eq:event1_2} or \eqref{eq:event2_2} or \eqref{eq:event4_2} is true} \}\right]\\
  &= \lfloor \max(v, 2s_{lim}(t))\rfloor+\sum_{t=\lfloor \min(v, 2s_{lim}(t)) \rfloor+1}^t \P\left(\text{\eqref{eq:event1_2} or \eqref{eq:event2_2} is true} \right)\\
 &\le \lfloor \max(v, 2s_{lim}(t))\rfloor+ 2\sum_{t=\lfloor \min(v, 2s_{lim}(t)) \rfloor+1}^t \frac{14}{t}
\end{align*}
using the harmonic series bound by $\log(t)+1$, we have
$$ \E[T_i(t)]\le \max(v, 2s_{lim}(t))+28(\log(t)+1)  $$

Then, we replace the value of $v$,

\textbf{First setting:} $\DeltaEpsi > 36\frac{\sigma_i^2}{\beta_i}\left(\sqrt{2} + 2\frac{\beta_i}{\sigma_i}\overline{\varepsilon} \right)^2$

$$ \E[T_i(t)]\le \log(t)\max\left(\frac{32\beta_i}{\DeltaEpsi}, \frac{8}{\left( p-5\varepsilon \right)^2}\left(1+2\sqrt{2}\left(\overline{\varepsilon}\vee \frac{9}{14\sqrt{2}} \right)\right)^2\right)+28(\log(t)+1)  $$

 \textbf{Second setting:} if $\DeltaEpsi \le  36\frac{\sigma_i^2}{\beta_i}\left(\sqrt{2} + 2\frac{\beta_i}{\sigma_i}\overline{\varepsilon} \right)^2$, then
$$\E[T_i(t)]\le \log(n)\max\left( \frac{40}{\DeltaEpsi^2}\left( \sigma_i\sqrt{2}+2\beta_i\overline{\varepsilon}\right)^2, \frac{8}{\left( p-5\varepsilon \right)^2}\left(1+2\sqrt{2}\left(\overline{\varepsilon}\vee \frac{9}{14\sqrt{2}} \right)\right)^2 \right)+28(\log(t)+1) . $$
}
\end{npf}
Finish the proof of the Theorem using the given values for the constants $\beta_i, \varepsilon, p$.

\subsection{Lemmas for concentration of robust estimators}
\subsubsection{Proof of Lemma~\ref{lem:bound_vpsi}: Controlling Variance of Influence of Huber's Estimator}
Let $\rho_\beta$ be Huber's loss function, with $\psi_\beta = \rho'_\beta$. We have that for any $x>0$, $\psi_\beta(x)^2\le 2\rho_\beta(x)$. Hence,
\begin{align*}
\Var(\psi_\beta(Y - \Hub_\beta(P)))&= \E[\psi_\beta(Y - \Hub_\beta(P))^2] \le 2\E[\rho_\beta(Y - \Hub_\beta(P))].
\end{align*}
Then, use that by definition of $\Hub_\beta(P)$, $\Hub_\beta(P)$ is a minimizer of $\theta \mapsto \E[\rho_\beta(Y - \theta)]$, hence,
\begin{align*}
\Var(\psi_\beta(Y - \Hub_\beta(P)))&\le 2\E[\rho_\beta(Y - \E[Y])].
\end{align*}
and finally, use that $\rho_\beta(x)\le x^2/2$ to conclude.

\subsubsection{Proof of Lemma~\ref{lem:tif_tT} : Concentrating Huber's Estimator by Concentrating the Influence}\label{sec:proof_tif_tT}
For all $n\in\N^*$, $\lambda >0$, let
$$f_{n}(\lambda)=\frac{\sign(\Delta_n)}{n}\sum_{i=1}^n \psi_\beta(X_i-\Hub_\beta(P)-\lambda \sign(\Delta_n)),$$
where $\Delta_n=  \Hub_\beta(P) - \Hub_\beta(X_1^n).$
\begin{npf}
	\item For any $\lambda>0$, $\P( |\Delta_n| \ge \lambda)\le \P(f_{n}(\lambda)\ge 0).\label{item:goal}
	$
	\\
	\onepf{
		For all $y \in \R$, let $J_n(y)=\frac{1}{n}\sum_{i=1}^n \rho_\beta(X_i-y)$  we have,
		\[
		J_n''(y) = \frac{1}{n}\sum_{i=1}^n \psi'_\beta\left(X_i-y\right).
		\]
		In particular, having $f_{n}(\lambda)=-\sign(\Delta_n)J'(\Hub_\beta(P) + \lambda \sign(\Delta_n))$ if we take the derivative of $f_{n}$ with respect to $\lambda$, we have the following equation
		\begin{align}\label{eq:deriv_neg}
		\frac{\partial}{\partial \lambda}f_{n}(\lambda)&=-\sign(\Delta_n)^2J_n''(\Hub_\beta(P) + \lambda \sign(\Delta_n) ) \nonumber\\
		&\le-\frac{1}{n}\sum_{i=1}^n\psi'_\beta(X_i-\Hub_\beta(P)-\lambda \sign(\Delta_n)).
		\end{align}
		Then, because $\psi'_\beta$ is non-negative, the function $\lambda \mapsto f_{n}(\lambda,)$ is non-increasing. Hence, for all $n\in\N^*$ and $\lambda>0$,
		$$ |\Delta_n| \ge \lambda \Rightarrow  f_{n}( | \Delta_n|)=0 \le f_{n}(\lambda),$$
		Hence,
		\begin{align}\label{eq:oracle1}
		\P( |\Delta_n| \ge \lambda)&\le \P( f_{n}(\lambda)\ge 0).\end{align}}
	\item For all $\lambda>0$, \label{item:taylor}
	\begin{equation*}
	f_{n}(\lambda)\le f_{n}(0)-\lambda\inf_{t \in [0,\lambda]}\left|f_{n}'(t)\right|. \end{equation*}\\
	\onepf{
		We apply Taylor's inequality to the function $f_{n}$.
		As $f_n$ is non-increasing (because its derivative is non-positive, see Equation~\eqref{eq:deriv_neg}), we get
		$$f_{n}(\lambda)\le f_{n}(0)-\lambda\inf_{t \in [0,\lambda]}\left|f_{n}'(t)\right|.$$
	}
	\item Let $m_n=\E\left[\inf_{t \in [0,\lambda]} \frac{1}{n}\sum_{i=1}^n \psi'_\beta(X_i'-\Hub_\beta(P)-t) \right]$. With probability larger than $1-2e^{-2n\eta^2}$, \label{item:stp3}
	$$\inf_{t \in [0,\lambda]}\left|f_{n}'(t))\right|\ge m_n-2\eta - \varepsilon,$$
	\onepf{
		Write that $X_i=(1-W_i)Y_i+W_i Z_i$ where $W_1,\dots, W_n$ are i.i.d Bernoulli random variable with mean $\varepsilon$, $Y_1,\dots, Y_n$ are i.i.d $\sim P$ and $Z_1,\dots, Z_n$ are i.i.d with law $H$.
		
		From equation~\eqref{eq:deriv_neg},
		\begin{align}\label{eq:oracle2}
		\left|f_{n}'(t))\right|\ge& \frac{1}{n}\sum_{i=1}^n \psi'_\beta(X_i-\Hub_\beta(P)-t\sign(\Delta))\nonumber\\
		\ge & \frac{1}{n}\sum_{i=1}^n \1\{W_i = 0\}\psi'_\beta(Y_i-\Hub_\beta(P)-t \sign(\Delta)) \\
		&+\frac{1}{n}\sum_{i=1}^n \1\{W_i = 1\}\psi'_\beta(Z_i-\Hub_\beta(P)-t \sign(\Delta))\\
		\ge & \frac{1}{n}\sum_{i=1}^n \psi'_\beta(Y_i-\Hub_\beta(P)-t \sign(\Delta))\\
		&+ \frac{1}{n}\sum_{i=1}^n \1\{W_i = 1\}\left(\psi'_\beta(Z_i-\Hub_\beta(P)-t \sign(\Delta))- \psi'_\beta(W_i-\Hub_\beta(P)-t \sign(\Delta))\right)
		\end{align}
		Hence,  because $\psi'_\beta \in [0,1]$, we have
		\begin{align}\label{eq:fndev}
		\left|f_{n}'(t))\right| &\ge \frac{1}{n}\sum_{i=1}^n \psi'_\beta(Y_i-\Hub_\beta(P)-t \sign(\Delta))- \frac{1}{n}\sum_{i=1}^n \1\{W_i = 1\})
		\end{align}
		The right-hand side depends on the infimum of the mean of $n$ i.i.d random variables in $[0,1]$. Hence, the function
		$$Z(X_1^n) \mapsto \sup_{t \in [0,\lambda]}\sum_{i=1}^n \psi'_\beta(X'_i-\Hub_\beta(P)-t) $$
		satisfies, by sub-linearity of the supremum operator and triangular inequality, the bounded difference property, with differences bounded by $1$. Hence, by Hoeffding's inequality, we get with probability larger than $1-e^{-2n\eta^2}$,
		\begin{align*}
		\inf_{t \in [0,\lambda]}\left|f_{n}'(t))\right|\ge& \E\left[\inf_{t \in [0,\lambda]} \frac{1}{n}\sum_{i=1}^n \psi'_\beta(X_i'-\Hub_\beta(P)-t) \right]-\eta- \frac{1}{n}\sum_{i=1}^n \1\{W_i = 1\})
		\end{align*}
		and using Hoeffding's inequality to control $ \frac{1}{n}\sum_{i=1}^n \1\{W_i = 1\}$, we have with probability larger than $1-2e^{-2\eta^2/n}$,
		\begin{align*}
		\inf_{t \in [0,\lambda]}\left|f_{n}'(t))\right|\ge& \E\left[\inf_{t \in [0,\lambda]} \frac{1}{n}\sum_{i=1}^n \psi'_\beta(X_i'-\Hub_\beta(P)-t) \right]-2\eta- \varepsilon
		\end{align*}
	}
	\item For $\lambda\in (0, \beta/2)$,
	$$\P\left(\quad|\Delta_n| \ge \lambda\right)\le    \P\left( \quad\left|\frac{1}{n}\sum_{i=1}^n \psi_\beta(X_i-\Hub_\beta(P))\right|\ge\lambda \left(m_n-\eta - \varepsilon \right)\right)+2e^{-2n\eta^2}.$$\\
	\onepf{
		For any $\lambda>0$, we have
		\begin{align}\label{eq:result_concentration}
		\P(|\Delta_n|  \ge \lambda)&\le \P(f_{n}(\lambda)\ge 0) && (\text{from \ref{item:goal}})\nonumber\\
		&\le 1-\P\left( f_{n}(0)-\lambda\inf_{t \in [0,\lambda]}\left|f_{n}'(t)\right|\le 0\right) && (\text{from \ref{item:taylor}})\nonumber\\
		&\le 1-\P\left( f_{n}(0)\le \lambda\left(m_n -2\eta- \varepsilon \right) \right)+2e^{-2n\eta^2} && (\text{from \ref{item:stp3}})\nonumber\\
		&=  \P\left( \left|\frac{1}{n}\sum_{i=1}^n \psi_\beta(X_i-\Hub_\beta(P))\right|\ge\lambda \left(m_n-\eta - \varepsilon \right)\right)+2e^{-2n\eta^2}.
		\end{align}
	}
	\item We prove that $m_n\ge p,$ and hence
	$$\P\left(  |\Delta_n| \ge \lambda\right)\le  \P\left(\left|\frac{1}{n}\sum_{i=1}^n \psi_\beta(X_i-\Hub(P))\right|\ge\lambda \left(p-\eta-\varepsilon\right)\right)+2e^{-2n\eta^2} $$
	\onepf{
		For all $\lambda \le \beta/2$,
		\begin{align*}
		\E\left[\inf_{\substack{t \in [0,\lambda]}}\frac{1}{n}\sum_{i=1}^n \psi'_\beta(X_i'-\Hub_\beta(P)-t) \right]&= \E\left[\inf_{t \in [0,\lambda]}\frac{1}{n}\sum_{i=1}^n\1\{|X_i'-\Hub_\beta(P)-t|\le \beta\} \right]\\
		&\ge \E\left[\frac{1}{n}\sum_{i=1}^n\1\{|X_i'-\Hub_\beta(P)|\le \beta-\lambda\} \right]\\
		&\ge \E\left[\frac{1}{n}\sum_{i=1}^n\1\{|X_i'-\Hub_\beta(P)|\le \beta/2\} \right]=p
		\end{align*}
		Then, we plug the bound on $m_n$ found in the previous step in equation \eqref{eq:result_concentration}, we get for any $\eta>0$ and $\lambda \in (0,\beta/2]$,
		\begin{align*}
		\P&( |\Delta_n| \ge \lambda)\le   \P\left( \left|\frac{1}{n}\sum_{i=1}^n \psi_\beta(X_i-\Hub_\beta(P))\right|\ge\lambda \left(p-\eta-\varepsilon\right)\right)+2e^{-2n\eta^2}
		\end{align*}
	}
\end{npf}

\subsubsection{Proof of Lemma~\ref{lem:polynomial_root_bound}: Algebra tool for bounding polinomial roots}
The solutions of the second order polynomial indicate that $x$ must verify
$$x \ge \frac{-b + \sqrt{b^2+4ac}}{2a} \ge \frac{b}{2a}\left( -1 + \sqrt{1+\frac{4ac}{b^2}}\right).$$
Then, use that the function $x \mapsto \sqrt{x+1}$ is concave and hence the graph of $x \mapsto \sqrt{x+1}$ is above its chords and we have for any $x \in [0,d]$, $\sqrt{1+x}\ge 1+x \frac{\sqrt{d+1}-1}{d}$. Hence,
$$x \ge \frac{b}{2a}\left( \frac{4ac(\sqrt{d+1}-1)}{db^2}\right)=\frac{2c(\sqrt{d+1}-1)}{db}.$$

\newpage
\subsubsection{Proof of Lemma~\ref{lem:integral_student}: Algebra on Student's distribution}\label{sec:proof_lem_integral_student}

 We have,
 \begin{align*}
  \int_\R \frac{\left(1+\frac{(y+a)^2}{d}\right)^{\frac{d+1}{2}}}{\left(1+\frac{y^2}{d}\right)^{d+1}}\mathrm{d}y& =\int_\R\sum_{l =0}^{\frac{d+1}{2}} {  \frac{d+1}{2} \choose l} \frac{(y+a)^{2l}}{d^{l}\left(1+\frac{y^2}{d}\right)^{d+1}}\mathrm{d}y\\
  &= \int_\R\sum_{l =0}^{\frac{d+1}{2}}\sum_{j=0}^{2l} {  \frac{d+1}{2} \choose l} { 2l \choose j} \frac{y^ja^{2l-j}}{d^{l}\left(1+\frac{y^2}{d}\right)^{d+1}}\mathrm{d}y\\
  &= \sum_{l =0}^{\frac{d+1}{2}}\sum_{j=0}^{2l} {  \frac{d+1}{2} \choose l} { 2l \choose j} \int_\R\frac{y^ja^{2l-j}}{d^{l}\left(1+\frac{y^2}{d}\right)^{d+1}}\mathrm{d}y
 \end{align*}
Remark that the integral is $0$ if $j$ is odd. Hence,
 \begin{align*}
  \int_\R \frac{\left(1+\frac{(y+a)^2}{d}\right)^{\frac{d+1}{2}}}{\left(1+\frac{y^2}{d}\right)^{d+1}}\mathrm{d}y& = \sum_{l =0}^{\frac{d+1}{2}}\sum_{j=1}^{l} {  \frac{d+1}{2} \choose l} { 2l \choose 2j} \frac{a^{2l-2j}}{d^{l}}\int_\R\frac{y^{2j}}{\left(1+\frac{y^2}{d}\right)^{d+1}}\mathrm{d}y
 \end{align*}
 Then, we compute the integrals. By change of variable $u=y/d$, we have
 \begin{align*}
  \int_\R\frac{y^{2j}}{\left(1+\frac{y^2}{d}\right)^{d+1}}\mathrm{d}y &=d^{j+1/2}\int_\R\frac{u^{2j}}{\left(1+u^2\right)^{d+1}}\mathrm{d}u\le  2d^{j+1/2}
 \end{align*}
 and for $l = j$,
 $$ \sum_{l=0}^{\frac{d+1}{2}}{ \frac{d+1}{2} \choose l} \frac{1}{d^{l}}\int_\R\frac{y^{2l}}{\left(1+\frac{y^2}{d}\right)^{d+1}}\mathrm{d}y = \int_\R \frac{(1+y^2/d)^{\frac{d+1}{2}}}{\left(1+\frac{y^2}{d}\right)^{d+1}} \mathrm{d}y$$
Hence,
 \begin{align}\label{eq:bound_integral_student}
  \int_\R \frac{\left(1+\frac{(y+a)^2}{d}\right)^{\frac{d+1}{2}}}{\left(1+\frac{y^2}{d}\right)^{d+1}}\mathrm{d}y& \le 2\sum_{l =1}^{\frac{d+1}{2}}\sum_{j=0}^{l-1} {  \frac{d+1}{2} \choose l }{ 2l \choose 2j} \frac{a^{2l-2j}}{d^{l}}d^{j+1/2}+\int_\R \frac{(1+y^2/d)^{\frac{d+1}{2}}}{\left(1+\frac{y^2}{d}\right)^{d+1}} \mathrm{d}y\nonumber\\
  &= 2\sum_{l =1}^{\frac{d+1}{2}} a^{2l}\sum_{j=0}^{l-1} {  \frac{d+1}{2} \choose l }{ 2l \choose 2j} \frac{a^{-2j}}{d^{l}}d^{j+1/2}+\int_\R \frac{(1+y^2/d)^{\frac{d+1}{2}}}{\left(1+\frac{y^2}{d}\right)^{d+1}} \mathrm{d}y\nonumber\\
  &\le 2\sum_{l =1}^{\frac{d+1}{2}} a^{2l}\sum_{j=0}^{l-1} {  \frac{d+1}{2} \choose l }{ 2l \choose 2j} \frac{a^{-2j}}{d^{l}}d^{j+1/2}+\int_\R \frac{(1+y^2/d)^{\frac{d+1}{2}}}{\left(1+\frac{y^2}{d}\right)^{d+1}} \mathrm{d}y\\
 \end{align}
And,

\begin{align*}
 \sum_{l =1}^{\frac{d+1}{2}} a^{2l}\sum_{j=0}^{l-1} {  \frac{d+1}{2} \choose l }{ 2l \choose 2j} \frac{a^{-2j}}{d^{l}}d^{j+1/2} &= \sqrt{d}\sum_{l =1}^{\frac{d+1}{2}} \sum_{j=0}^{l-1} {  \frac{d+1}{2} \choose l }{ 2l \choose 2j} a^{2(l-j)}d^{j-l}\\
 &\le  \sqrt{d}\sum_{l =1}^{\frac{d+1}{2}} \sum_{j=0}^{l-1} {  \frac{d+1}{2} \choose l }{ 2(l-1) \choose 2j} l^2 \left(\frac{a^2}{d}\right)^{l-j}.
 \end{align*}
Using that ${ 2l \choose 2j}={ 2(l-1) \choose 2j}\frac{2l(2l-1)}{(2l-2j)(2l-2j-1)}\le { 2(l-1) \choose 2j}l^2$.\\
Then, completing the binomial sum so that 
$$ \sum_{j=0}^{l-1}{ 2(l-1) \choose 2j} \left(\frac{a^2}{d}\right)^{-j}\le \sum_{j=0}^{2(l-1)}{ 2(l-1) \choose 2j} \left(\frac{a^2}{d}\right)^{-j}=\left(1+\frac{\sqrt{d}}{a}\right)^{2(l-1)},  $$
we have,
\begin{align*}
  \sum_{l =1}^{\frac{d+1}{2}} a^{2l}\sum_{j=0}^{l-1} {  \frac{d+1}{2} \choose l }{ 2l \choose 2j} \frac{a^{-2j}}{d^{l}}d^{j+1/2}&\le \frac{1}{2}(d+1)\sqrt{d}\sum_{l =1}^{\frac{d+1}{2}}  {  \frac{d+1}{2} \choose l }  \left(\frac{a^2}{d}\right)^{l} l \left(1+\frac{\sqrt{d}}{a} \right)^{2(l-1)}\\
 &=  \frac{a^2}{2d}(d+1)\sqrt{d}\sum_{l =1}^{\frac{d+1}{2}}  {  \frac{d+1}{2} \choose l }  l  \left(\frac{a}{\sqrt{d}}+1 \right)^{2(l-1)}\\
 &=  \frac{a^2}{2d}(d+1)\sqrt{d}\sum_{l =0}^{\frac{d-1}{2}}  {  \frac{d-1}{2} \choose l } \frac{(d+1)(l+1)}{2(l+1)}  \left(\frac{a}{\sqrt{d}}+1 \right)^{2l}\\
 &\le   \frac{a^2}{4\sqrt{d}}(d+1)^2   \left(2+ \frac{a}{\sqrt{d}} \right)^{d-1}
\end{align*}

Then, inject this in Equation~\eqref{eq:bound_integral_student} to get
 \begin{align*}
  \int_\R \frac{\left(1+\frac{(y+a)^2}{d}\right)^{\frac{d+1}{2}}}{\left(1+\frac{y^2}{d}\right)^{d+1}}\mathrm{d}y& \le \frac{a^2}{2\sqrt{d}}(d+1)^2   \left(2+ \frac{a}{\sqrt{d}} \right)^{d-1}+\int_\R \frac{(1+y^2/d)^{\frac{d+1}{2}}}{\left(1+\frac{y^2}{d}\right)^{d+1}} \mathrm{d}y.
 \end{align*}

\clearpage
\section{Additional experimental results}

\subsection{Sensitivity to $\beta$ and $\varepsilon$}\label{sec:choice_param}

In this section we illustrate the impact of the choice of $\beta$ and $\varepsilon$ on the estimation.

\begin{figure}[h!]
	\caption{Cumulative regret plots for different values of the parameters $\varepsilon$ and $\beta$ on a Weibull dataset.\label{fig:corrupted_weibull}}
	\subfigure[Dependency in $\varepsilon$\label{fig:dep_epsi}]{
		\includegraphics[width=0.5\textwidth]{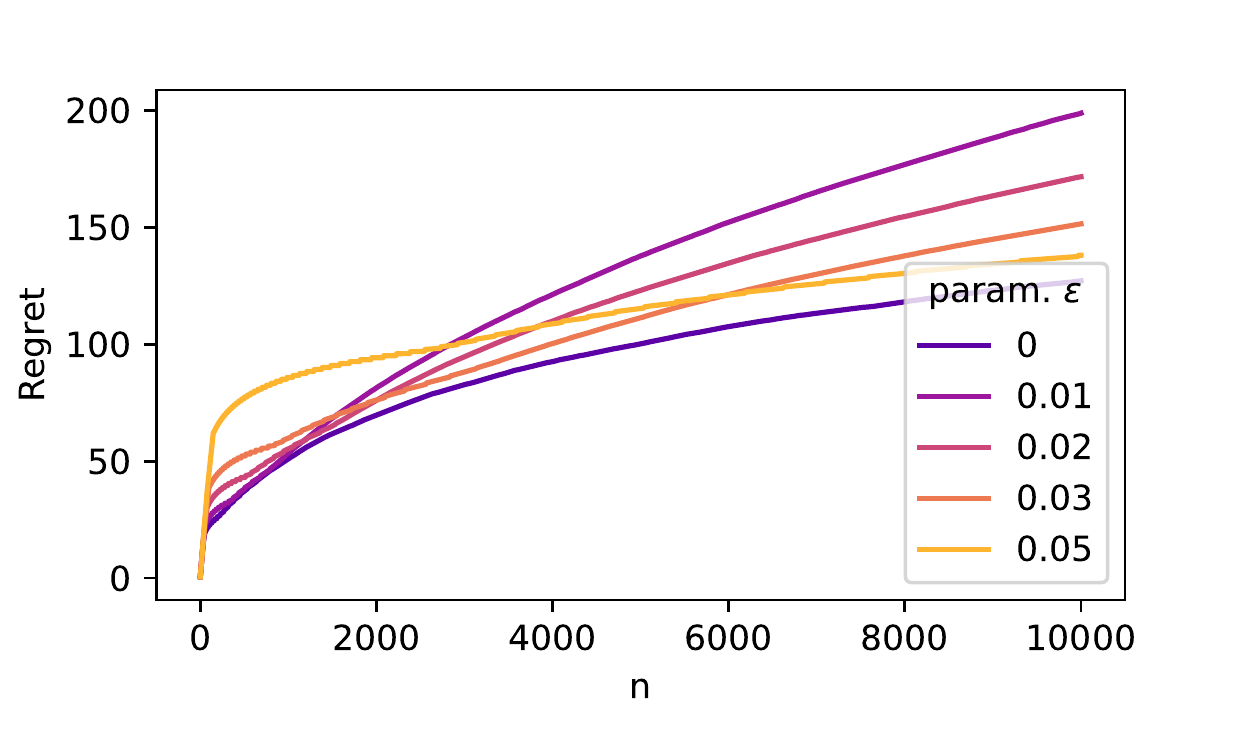}}
	\subfigure[Dependency in $\beta$\label{fig:dep_beta}]{
		\includegraphics[width=0.5\textwidth]{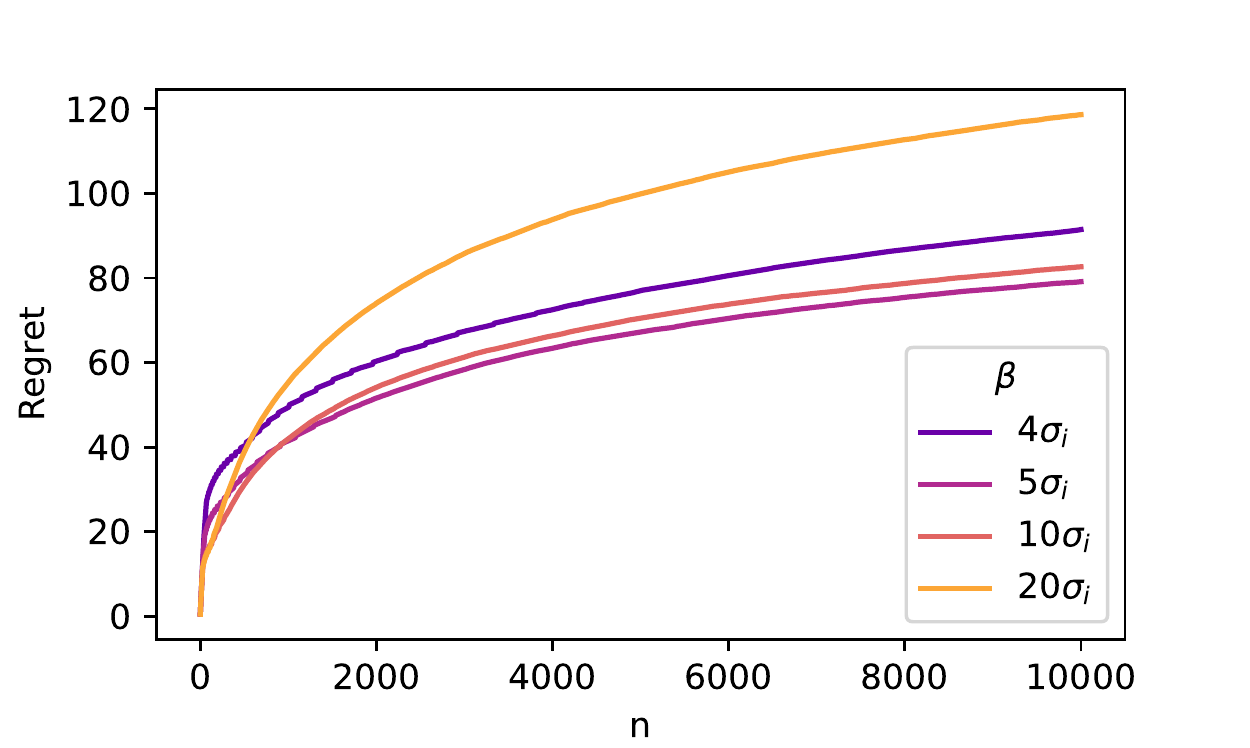}
	}
\end{figure}

\paragraph{Choice of $\beta$ (Figure~\ref{fig:dep_beta}):}
The choice of $\beta$ is a trade-off between the bias (distance $|\Hub_\beta(P)-\E[X]|$ which decreases as $\beta$ go to infinity) and robustness (when $\beta$ goes to $0$, $\Hub_\beta(P)$ goes to the median). To illustrate this trade-off we use the Weibull distribution for which can be very asymmetric. We use a 3-armed bandit problem with shape parameters $(2, 2, 0.75)$ and scale parameters $(0.5, 0.7, 0.8)$ which implies that the means are approximately $(0.44, 0.62, 0.95)$. These distributions are very asymmetric, hence the bias $|\Hub_\beta(P)-\E[X]|$ is high and in fact even though arm 3 has the optimal mean, arm 2 will have the optimal median, the medians are given by $(0.41, 0.58,0.49)$. In this experiment we don't use any corruption as we don't want to complicate the interpretation. As expected by the theory, we get that $\beta_i$ should not be too small or too large but it should be around $4\sigma_i$.

\paragraph{Choice of $\varepsilon$ (Figure~\ref{fig:dep_epsi}):}
To illustrate the dependency in $\varepsilon$, we also use the Weibull distribution to show the dependency in $\varepsilon$ with the same parameters as in the previous Weibull example, except that we choose $\beta_i = 5\sigma_i$ which is around the optimum found in the previous experiment and we corrupt with $2\%$ of outliers (this is the true $\varepsilon$ while we will make the $\varepsilon$ used in the definition of the algorithm vary). The outliers are constructed as in Section~\ref{sec:xp1}. The effect of the parameter $\varepsilon$ is difficult to assess because $\varepsilon$ has an impact on the length of force exploration that we impose at the beginning of our algorithm (the $s_{lim}$).

\end{document}